\documentclass[screen]{acmart}

\usepackage{booktabs} 

\usepackage{graphicx}  
\usepackage{amsfonts} 
\usepackage{CJK}
\usepackage{amsmath}
\usepackage{bm}
\usepackage{multirow}
\usepackage{subfigure}
\usepackage{makecell}
\usepackage{xspace}
\usepackage{array}
\usepackage{colortbl}
\usepackage{soul}
\usepackage{url}
\usepackage{amsmath}
\usepackage{subfigure}
\usepackage[utf8]{inputenc}
\usepackage{graphicx}
\usepackage{amsmath}
\usepackage{booktabs}
\usepackage{xcolor}
\newcommand{\hlc}[2][yellow]{{%
		\colorlet{foo}{#1}%
		\sethlcolor{foo}\hl{#2}}%
}
\usepackage{algorithm}
\usepackage{algorithmic}
\usepackage{colortbl}
\usepackage[misc]{ifsym}
\usepackage{makecell}
\usepackage{color}

\newcommand{\ie}{\emph{i.e.,}\xspace}

\newcommand{\eg}{\emph{e.g.,}\xspace}

\newcommand{\ignore}[1]{}

\newcommand{\dubbelop}{$^{\blacktriangle}$}

\newcommand{\dubbelneer}{$^{\blacktriangledown}$}

\AtBeginDocument{%
  \providecommand\BibTeX{{%
    \normalfont B\kern-0.5em{\scshape i\kern-0.25em b}\kern-0.8em\TeX}}}

\setcopyright{acmcopyright}
\acmJournal{TOIS}
\acmYear{2022} \acmVolume{1} \acmNumber{1} \acmArticle{1} \acmMonth{1} \acmPrice{15.00}\acmDOI{10.1145/3517221}

\title{Follow the Timeline! Generating Abstractive and Extractive Timeline Summary in Chronological Order}

\begin{document}

\author{Xiuying Chen}
\authornotemark[1]
\affiliation{%
  \institution{Computational Bioscience Reseach Center, King Abdullah University of Science and Technology}
}
\email{xiuying.chen@kaust.edu.sa}

\author{Mingzhe Li}
\authornote{Equal contribution. Ordering is decided by a coin flip.}
\affiliation{%
  \institution{Wangxuan Institute of Computer Technology, Peking University}
}
\email{li_mingzhe@pku.edu.cn}

\author{Shen Gao}   
\affiliation{%
  \institution{Wangxuan Institute of Computer Technology, Peking University}
}
\email{shengao@pku.edu.cn}

\author{Zhangming Chan} 
\affiliation{%
  \institution{Wangxuan Institute of Computer Technology, Peking University}
}
\email{zhangming.chan@pku.edu.cn}

\author{Dongyan Zhao}
\affiliation{%
    \institution{Wangxuan Institute of Computer Technology, Peking University}
}
\email{zhaody@pku.edu.cn}

\author{Xin Gao}
\affiliation{%
    \institution{Computational Bioscience Reseach Center, King Abdullah University of Science and Technology}
}
\email{xin.gao@kaust.edu.sa}

\author{Xiangliang Zhang}
\affiliation{%
    \institution{\textsuperscript{1} University of Notre Dame; 
  \textsuperscript{2} King Abdullah University of Science and Technology}
}
\email{xzhang33@nd.edu}

\author{Rui Yan}
\authornote{Corresponding Author: Rui Yan (ruiyan@ruc.edu.cn)}
\affiliation{
  \institution{ Gaoling School of Artificial Intelligence, Renmin University of China}
}
\email{ruiyan@ruc.edu.cn}

\begin{abstract}
Nowadays, time-stamped web documents related to a general news query floods spread throughout the Internet, and timeline summarization targets concisely summarizing the evolution trajectory of events along the timeline.
Unlike traditional document summarization, timeline summarization needs to model the time series information of the input events and summarize important events in chronological order.
To tackle this challenge, in this paper, we propose a \emph{Unified Timeline Summarizer} (UTS) that can generate abstractive and extractive timeline summaries in time order.
Concretely, in the encoder part, we propose a graph-based event encoder that relates multiple events according to their content dependency and learns a global representation of each event. 
In the decoder part, to ensure the chronological order of the abstractive summary, we propose to extract the feature of event-level attention in its generation process with sequential information remained and use it to simulate the evolutionary attention of the ground truth summary.
The event-level attention can also be used to assist in extracting summary, where the extracted summary also comes in time sequence.
We augment the previous Chinese large-scale timeline summarization dataset and collect a new English timeline dataset.
Extensive experiments conducted on these datasets and on the out-of-domain Timeline 17 dataset show that UTS achieves state-of-the-art performance in terms of both automatic and human evaluations\footnote{\url{https://github.com/iriscxy/Unified-Timeline-Summarizer}}.

\end{abstract}



\maketitle

\section{Introduction}
	\label{sec:intro}
	The rapid growth of World Wide Web means that time-stamped document floods spread throughout the Internet. 
	General search engines simply return web pages ranked by query relevance, but they are not quite capable of handling ambiguous intentioned queries, such as a query about evolving news ``COVID-19''. 
	People may have a myriad of general interests about the beginning, the evolution, or the most up-to-date situation, while simply ranking the returned webpages according to their relevance is insufficient.
	In many cases, readers are tired of navigating every document in the overwhelming collection: they want to monitor the evolution trajectory of hot topics by simply browsing.
	Summarization is an ideal solution to provide a condensed, informative document reorganization for a faster and better representation of news evolution. 
	Timeline summary temporally summarizes evolutionary news as a series of individual but correlated component summaries and hence offers an option to understand the big picture of a developing situation~\citep{yan2011evolutionary}.
	
	Existing timeline summarization approaches such as \citep{yan2011evolutionary,li2013evolutionary,ren2013personalized} are all based on extraction methods.
	However, these methods rely on human-engineered features and sophisticated abilities that are crucial to high-quality summarization, such as paraphrasing, generalization, or the incorporation of real-world knowledge, which are possible only in an abstractive framework.
	Recently, with the emergence of strong generative neural models for text \citep{Bahdanau2015Neural}, abstractive techniques are also becoming increasingly popular.
	Hence, we propose the \emph{abstractive timeline summarization task} in our early work~\citep{Chen2019Learning}, which aims to concisely paraphrase the event information in the input article.
	An example case is shown in Table~\ref{tab:intro-case}, where the article consists of events of a great entertainer in different periods, and the summary correctly summarizes the important events from the input article in order.

	Abstractive summarization approaches including \cite{getto17,unified18,gavrilov2019self,zhang2020structure} have been proven to be useful in traditional summarization task. 
	However, unlike traditional document summarization, the timeline summarization dataset consists of a series of time-stamped events, and it is crucial for the timeline summarization model to capture this time series information to better guide the chronological summary generation process.
	Besides, the fidelity problem is also of vital importance for timeline summarization, where mixing the information of different events leads to a bad summary.
	Take the example in Table~\ref{tab:intro-case} for example, the bad summary confuses the birthplace and the residence, the first album, and the best-selling album of the celebrity.
	Herein, the good summary is the ground truth summary from our dataset, and the bad summary is a wrong summary with typical errors we found in a preliminary experiment.
	As we found in the experiment, such infidelity phenomena is a commonly-faced problem in summarization tasks.

	To tackle the above challenges, in our previous work~\citep{Chen2019Learning}, we come up with a \emph{Memory-based Timeline Summarization} (MTS) model.
	Specifically, we first use an event embedding module with selective reading units to embed all events.
	Then, we propose a key-value memory module storing time-series information to guide the summary generation process.
	Concretely speaking, the key in the memory module is the time position embedding that represents the time series information, and the values are the corresponding event representations.
	The value item includes local and global representation, where local value is the output from the event embedding module, and global value is taken from the average local representation.
	Keys together form a timeline and we use the time position of events on the timeline to guide the generation process.
	Finally, in each decoding step, we introduce event-level attention and use it to determine word-level attention to avoid confusion between events.

\begin{table}[]
	\centering
	\begin{tabular}{l|l}
		\toprule
		Events & \makecell*[c]{\multicolumn{1}{p{12cm}}{Michael Jackson (dubbed as ``King of Pop'') was born on  \hlc[pink!50]{August 29, 1958} in Gary, Indiana. He is the seventh child in his family.
		}}       \\
		\cline{2-2} & \makecell*[c]{\multicolumn{1}{p{12cm}}
			{In \hlc[pink!50]{1971}, Jackson released his first solo ``got to be there'', marking the beginning of his solo career.}} \\
		\cline{2-2} & \makecell*[c]{\multicolumn{1}{p{12cm}}
			{In \hlc[pink!50]{late 1982}, Jackson's sixth album, ``Thriller'', was released, where videos "Beat It", "Billie Jean" in it are credited with breaking racial barriers and transforming the medium into an art form and promotional tool.}} \\
		\cline{2-2} & \makecell*[c]{\multicolumn{1}{p{12cm}}
			{In \hlc[pink!50]{March 1988}, Jackson built a new home named Neverland Ranch in California, where more than 100 arcade machines were stored here.}} \\
		\cline{2-2} & \makecell*[c]{\multicolumn{1}{p{12cm}}
			{In \hlc[pink!50]{2000}, Guinness World Records recognized him for supporting 39 charities and donated more than 300 million dollars to charities in his own name, more than any other entertainer. }} \\
		\hline
		Bad summary & \makecell*[c]{\multicolumn{1}{p{12cm}}
			{Michael Jackson was born on  \hlc[pink!50]{August 29, 1958} in \hlc[cyan!30]{Gary, California}.
				In \hlc[pink!50]{1971},  \hlc[cyan!30]{his first album ``Thriller''} was released.
				In \hlc[pink!50]{2000}, Guinness World Records recognized him for supporting 39 charities.
		}} \\ \hline
		
		\multicolumn{1}{r|}{Good summary}& \makecell*[c]{\multicolumn{1}{p{12cm}}
			{Michael Jackson was born on  \hlc[pink!50]{August 29, 1958} in \hlc[pink!50]{Gary, Indiana}. 
	    \hlc[pink!50]{	His sixth album ``Thriller''} was released in \hlc[pink!50]{1982}.
				In \hlc[pink!50]{2000}, Guinness World Records recognized him for supporting 39 charities.
		}}
		\\ 
		\bottomrule
	\end{tabular}
		\caption{Example of timeline summarization. The text in pink demonstrates time stamp, and text in blue demonstrates wrong event description. Events are split by lines.}
	\label{tab:intro-case}
\end{table}

	In MTS, the time information is captured in an implicit and indirect way. 
	MTS stores the time position embedded in the memory and hopes the decoder will learn to attend to the correct time position in the training process.
	However, that strategy is rather weak supervision, where it is hard to verify and ensure the decoder indeed captures the time-sequential information.
	In this work, we take one step further and improve our previously proposed MTS framework with explicit timeline guidance modeling. 
	In other words, we carefully design a strategy that lets the time information be a clear guidance signal for the summarization process.	
	
	Overall, in this paper, we propose a novel \emph{Unified Timeline Summarizer} (UTS) that can generate abstractive and extractive timeline summaries in time order.
	For the abstractive part, concretely, in the encoder part, we first propose a graph-based event encoder that relates multiple events according to their content dependency and learns a representation of each event. 
	The motivation is that the importance of each event and whether it should be included in the summary does not only depend on itself but also is related to other events.
	Take Table~\ref{tab:intro-case} for example, Jackson releases his first solo album might be an important event, but its importance is weakened by his ``Thriller'' album that breaks the racial barriers.
	Hence, the representation from the graph encoder incorporates global information from other events, thus is used to replace the old global representation in the memory.
	In the decoder part, to avoid the situation in the bad summary in Table~\ref{tab:intro-case}, where it confuses the birthplace and the residence because the model is not sensitive to the timeline, we propose a summary decoder that emphasizes the time information.
	Concretely, to ensure the chronological order of the abstractive summary, we propose to extract the feature of event-level attention in its generation process with sequential information remained and use it to simulate the evolutionary attention of the ground truth summary.
	
	In terms of the extractive part, we present a sentence embedding module to encode each sentence.
	Next, a sentence extractor sequential selects important sentences to be included in the summary.
	The event-level attention can also be used to assist in extracting summary in this process, where we devise a time-aware inconsistency loss function to penalize the inconsistency between abstractive attention and extractive attention.
	Note that the extractive summary is extracted one by one, thus the extracted summary also comes in time sequence.

	We empirically compare MTS and UTS on the public dataset\footnote{\url{https://github.com/yingtaomj/Learning-towards-Abstractive-Timeline-Summarization}} proposed by our early work~\citep{Chen2019Learning}. 
	This is a large-scale real-world timeline summarization dataset, which consists of a series of time-stamped events and the corresponding summary.
	Moreover, since this previous dataset only includes a timeline corpus about celebrities, we augment the dataset with cases about social events.
	We also collect an English timeline summarization dataset.
	Experimental results on these datasets and on out-of-domain Timeline17 dataset show that our newly proposed UTS model can significantly outperform the existing methods. 
	Particularly, UTS-\textit{abs} yields 4.47\% and 5.90\% percentage point improvement in terms of ROUGE-1 on celebrity and event timeline datasets compared with our early work MTS. 
	In addition to the comprehensive evaluation, we also evaluate our proposed graph encoder and attention mechanism by a fine-grained analysis. 
	The analysis reveals how the model leverages the explicit timeline information to guide the abstractive and extractive summarization process and provides us insights on why they can achieve big improvement over state-of-the-art methods.

	Overall, our contributions can be summarized as follows:

	$\bullet$ We propose a unified abstractive and extractive timeline summarization framework, where a time-aware inconsistency loss function is proposed to unify these two processes.
	
	$\bullet$ We propose a graph-based encoder that relates multiple events according to their content dependency and learns the global representation of each event.
	
	$\bullet$ We propose to use the evolutionary attention of the ground truth summary to guide both the abstractive and extractive summary generation process, to ensure that the generated summaries follow strict time order.
	
	$\bullet$  We also augment the first real-world large-scale timeline summarization dataset with social event corpus and corpus in English\footnote{Data will be released in camera-ready version.}.
	Experiments conducted on the three datasets and the out-of-domain benchmark Timeline 17 dataset show that our model outperforms all baselines, including state-of-the-art models. 
	Experiments also verify the effectiveness of each module in UTS as well as its interpretability.
	
	The rest of the paper is organized as follows: We summarize related work in \S\ref{sec:related}. 
	We then formulate our research problem in \S\ref{sec:Problem} and elaborate our approach in \S\ref{sec:model}. 
	\S\ref{sec:setup} gives the details of our experimental setup and \S\ref{sec:results} presents the experimental results. 
	Finally, \S\ref{sec:conclusion} concludes the paper.

\section{Related Work}
	\label{sec:related}
	
	We detail related work on text generation methods, timeline summarization, extractive summarization, abstractive summarization, unified summarization, and memory network.
	
	\subsection{Text Generation Methods}
	
In recent years, sequence-to-sequence (seq2seq)~\cite{Sutskever2014SequenceTS} based neural networks have been proved effective in generating a fluent sentence.
The seq2seq model is originally proposed for machine translation and later adapted to various natural language generation tasks, such as text summarization~\cite{Wang2019Concept,Paulus2018A,Gehrmann2018BottomUp,Lin2018Global,Wang2019BiSET} and dialogue generation~\cite{Tao2018RUBERAU,Yao2017TowardsIC,Cai2019Retrievalguided,Zhao2020LowResource,Zhang2020Modeling}.
\citet{rush2015neural} apply the seq2seq mechanism with attention model to the text summarization field.
Then \citet{getto17} add copy mechanism and coverage loss to generate summarization without out-of-vocabulary and redundancy words.
The seq2seq architecture has also been broadly used in a dialogue system.
\citet{Tao2018Get} propose a multi-head attention mechanism to capture multiple semantic aspects of the query and generate a more informative response.
\citet{Yao2017TowardsIC} propose to use the content introducing method to solve the problem of generating a meaningless response.
\citet{Wang2018Chat} use three channels for widening and deepening the topics of interest and try to make the conversational model chat more turns.
	
	\subsection{Timeline Summarization}
	The timeline summarization task is firstly proposed by \citet{allan2001temporal}, where they define temporal summaries of news stories as extracting a single sentence from each event within a news topic.
	Later, a series of works \citep{yan2011evolutionary,yan2011timeline,yan2012visualizing,zhao2013timeline} further investigate timeline summarization task.
	\citet{yan2011evolutionary} formally formulate the task as an optimization problem via iterative substitution from a set of sentences to a subset of sentences that satisfies the above requirements, balancing coherence/diversity measurement and local/global summary quality.
	In follow-up work, \citet{yan2011timeline} propose to model trans-temporal correlations among component summaries for timelines, using inter-date and intra-date sentence dependencies, and present a novel combination.
	There are also works focusing on tweets summarization that is related to timeline summarization.
	For example, \cite{ren2013personalized} focus on the problem of selecting meaningful tweets given a user's interests; the dynamic nature of user interests, the sheer volume, and the sparseness of individual messages make this a challenging problem.  
	Specifically, they consider the task of time-aware tweets summarization, based on a user’s history and collaborative social influences from ``social circles''.
	\citet{Ghalandari2020ExaminingTS} compare different timeline summarization strategies using appropriate evaluation frameworks.
	For a more robust evaluation, they also present a new timeline summarization dataset, which spans longer time periods than previous datasets.
	However, all the above works are based on extractive methods, which are not as flexible as abstractive approaches.
	
	The most similar work to ours is proposed by \citep{steen2019abstractive}, where they construct a word-adjacency graph, and then generate new sentences from this graph by finding paths from the sentence start node to the sentence end node.
	This is very different from our neural-based approach, and we demonstrate the superiority of our model in the experiment.

	\subsection{Extractive Summarization}
	
	Despite the focus on abstractive summarization, extractive summarization remains an attractive method.
	In extractive summarization,
    \citet{kobayashi2015summarization} propose a summarization method using document-level similarity based on word embeddings.
    Meanwhile, \citet{filippova2015sentence} use an RNN to delete words in a document for sentence compression.
     \citet{Yan2015DeepDS} propose more meaningful and informative units named frequent deep dependency sub-structure and a topic-sensitive multi-task learning model for multi-doc summarization.
    \citet{Cheng2016Neural} propose a general framework for single-document text summarization using a hierarchical article encoder composed with an attention-based extractor.
    Following this, \citet{nallapati2017summarunner} propose a simple RNN-based sequence classifier that outperforms or matches the state-of-art models at the time.
	\citet{chen2018iterative} introduce a model which iteratively polishes the document representation on many passes through the document, so as to produce a better summary.
	In another approach, \citet{Narayan2018Ranking} use a reinforcement learning method to optimize the ROUGE evaluation metric for text summarization.
	\citet{Ren2018SentenceRF} study the use of sentence relations, \eg contextual sentence relations, title sentence relations, and query sentence relations, so as to improve the performance of extractive summarization. 
	
	Recently, pre-trained language models are also applied in summarization for contextual word representations \cite{zhong2020extractive,liu2019text}.
	Another intuitive structure for extractive summarization is the graph, which can better utilize the statistical or linguistic information between sentences. 
	Early works focus on document graphs constructed with the content similarity among sentences, like LexRank \cite{erkan2004lexrank} and TextRank \cite{mihalcea2004textrank}. 
	Some recent works aim to incorporate a relational prior into the encoder by graph neural networks (GNNs)  \cite{yasunaga2017graph}. 
	

		\subsection{Abstractive Summarization} 
	Recently, with the emergence of strong generative neural models for text \citep{bahdanau2014neural}, abstractive summarization is also becoming increasingly popular \citep{nallapati2017summarunner,getto17}.
	These models typically take the form of convolutional neural networks (CNN) or recurrent neural networks (RNN).
For example, \citet{rush2015neural} propose an encoder-decoder model which uses a local attention mechanism to generate summaries.
\citet{Nallapati2016Abstractive} further develop this work by addressing problems that had not been adequately solved by the basic architecture, such as keyword modeling and capturing the hierarchy of sentence-to-word structures.
In follow-up work, \citet{Nallapati2017SenGen} propose a new summarization model which generates summaries by sampling a topic one sentence at a time, then producing words using an RNN decoder conditioned on the sentence topic. 
\citet{Zhu2020AttendTA} tackles the cross-lingual summarization task, which aims at summarizing a document in one language into another language. 
They propose a method inspired by the translation pattern in the process of obtaining a cross-lingual summary. 
A series of works relies on prototype text to assist in summarization. 
\citet{cao2018retrieve} chose the template with the highest similarity to the input sentence as a soft template to generate summaries.
Following this, \citet{gao2019write} proposed to generate the summary with pattern based on prototype editing.
Summarization techniques have also been used in other tasks such as related work generation \cite{chen2021capturing} and headline generation \cite{li2021style}.

        \subsection{Unified Summarization}
        Unified summarization here means unifying extractive and abstractive summarization tasks together.
        It is a common way to propose a multi-task framework that utilizes the benefits from one task to augment the performance of the other task.
        For example, 
\citet{unified18} proposed a unified framework that takes advantage of both extractive and abstractive summarization using an attention mechanism, which is a combination of the sentence-level attention.
\citet{Chen2018FastAS} introduced a multi-step procedure, namely compression paraphrase, for abstractive summarization, which first extracts salient sentences from documents and then rewrites them in order to get final summaries.
\citet{Li2018GuidingGF} introduced a guiding generation model, where the keywords in source texts are first retrieved with an extractive model.
	The most similar work to ours is \cite{unified18}, where they use sentence-level attention to modulate the word-level attention such that words in less attended sentences are less likely to be generated.
	Their sentence-level attention is static during the generation process, while in our model, the high-level attention changes in each decode step depending on the current generated word which is more reasonable.

	\subsection{Memory Network}
The memory network proposed by \citet{Sukhbaatar2015EndToEndMN} generally consists of two components.
The first one is a memory matrix to save information (\ie memory slots) and the second one is a neural network to read/write the memory slots.
The memory network has shown better performance than traditional long-short term memory network in several tasks, such as question answering~\cite{Sukhbaatar2015EndToEndMN,Pavez2018Working,Ma2018Visual,Gao2018MotionAppearance}, machine translation~\cite{Maruf2018Document}, text summarization~\cite{Kim2019Abstractive,Chen2019Learning}, dialog system~\cite{Chu2018Learning,Wu2019Globaltolocal}, job-resume matching~\cite{yan2019interview} and recommendation~\cite{Ebesu2018Collaborative,Wang2018Neural,Zhou2019TopicEnhanced}.
The reason is that the memory network can store the information in a long time range and has more memory storage units than LSTM which has a single hidden state.
Following memory network, there are many variations of memory network have been proposed, \ie key-value memory network~\cite{Miller2016KeyValueMN} and dynamic memory network~\cite{Xiong2016DynamicMN,Kumar2016AskMA}.
Representative works include \cite{gao2020meaningful}, where they generate more meaningful answers in E-commerce question-answering by a read-and-write memory consisting of selective writing units to conduct reasoning among these reviews.

	In our work, we apply the key-value memory network on the timeline summarization task and fuse it into the generation process.

	\section{Problem Formulation}
	\label{sec:Problem}

Before detailing our answer generation model, we first introduce our notations listed in Table~\ref{tbl:notations}.

\begin{table}[!t]
 \centering
 \begin{tabular}{ll}
  \toprule
  Symbol & Description \\
  \midrule 
  $X$ & a document consists of multiple events  \\
  $Y$ & ground truth timeline summary \\
  $\hat{Y}$ & generated timeline summary\\
  $x_i$ & $i$-th event in input document \\
  $w^i_j$ & $j$-th word in $i$-th event\\
  $T_e$ & number of input events \\
  $T_{w}^i$ & number of words in $i$-th event \\
  $T_y$ & number of words in ground truth summary \\
  $T_{ys}$ & number of sentences in the ground truth timeline summary
    \\
    $l_i$ &extract label for $i$-th sentence in the summary\\
  \bottomrule
 \end{tabular}
 \caption{Glossary.}
 \label{tbl:notations}
\end{table}

	UTS takes a list of events $X=(x_{1},...,x_{T_{e}})$ as inputs, where $T_{e}$ is the number of events.
	Each event $x_{i}$ is a list of words: $x_{i}= (w_{1}^i,w_{2}^i,...,w_{T_{w}^i}^i)$, where $w_j^i$ is the $j$-th word in $i$-th event, and $T_w^i$ is the word number of event $x_i$.
	
	In the abstractive part, UTS-\textit{abs} aims to generate a summary $\hat{Y}=(\hat{y}_1,...,\hat{y}_{T_y})$ that is not only grammatically correct but also consistent with the event information such as occurrence place and time.
	Essentially, UTS-\textit{abs} tries to optimize the parameters to maximize the probability $P(Y|X) = \prod_{t=1}^{T_y} P(y_t|X)$, where $Y=(y_1,...,y_{T_y})$ is the ground truth summary.
	
	For the extractive part, 
	UTS-\textit{ext} targets at generating a score vector $\hat{L} = \{\hat{l}_{1},\dots,\hat{l}_{T_{ys}}\}$ for each sentence, where each score denotes the sentence's extracting probability.
	We convert the human-written summaries to gold label vector  $L=\{l_{1},...,l_{T_{ys}}\}$, where $l_{i}\in \{0,1\}$ denotes whether the $i$-th sentence is selected (1) or not (0).
	During the training process, the cross-entropy loss is calculated between $L$ and $\hat{L}$, which is minimized to optimize $\hat{L}$.
	
	\section{Model}
	\label{sec:model}
	
	\subsection{Overview}
\begin{figure*}
		\centering
		\includegraphics[scale=0.43]{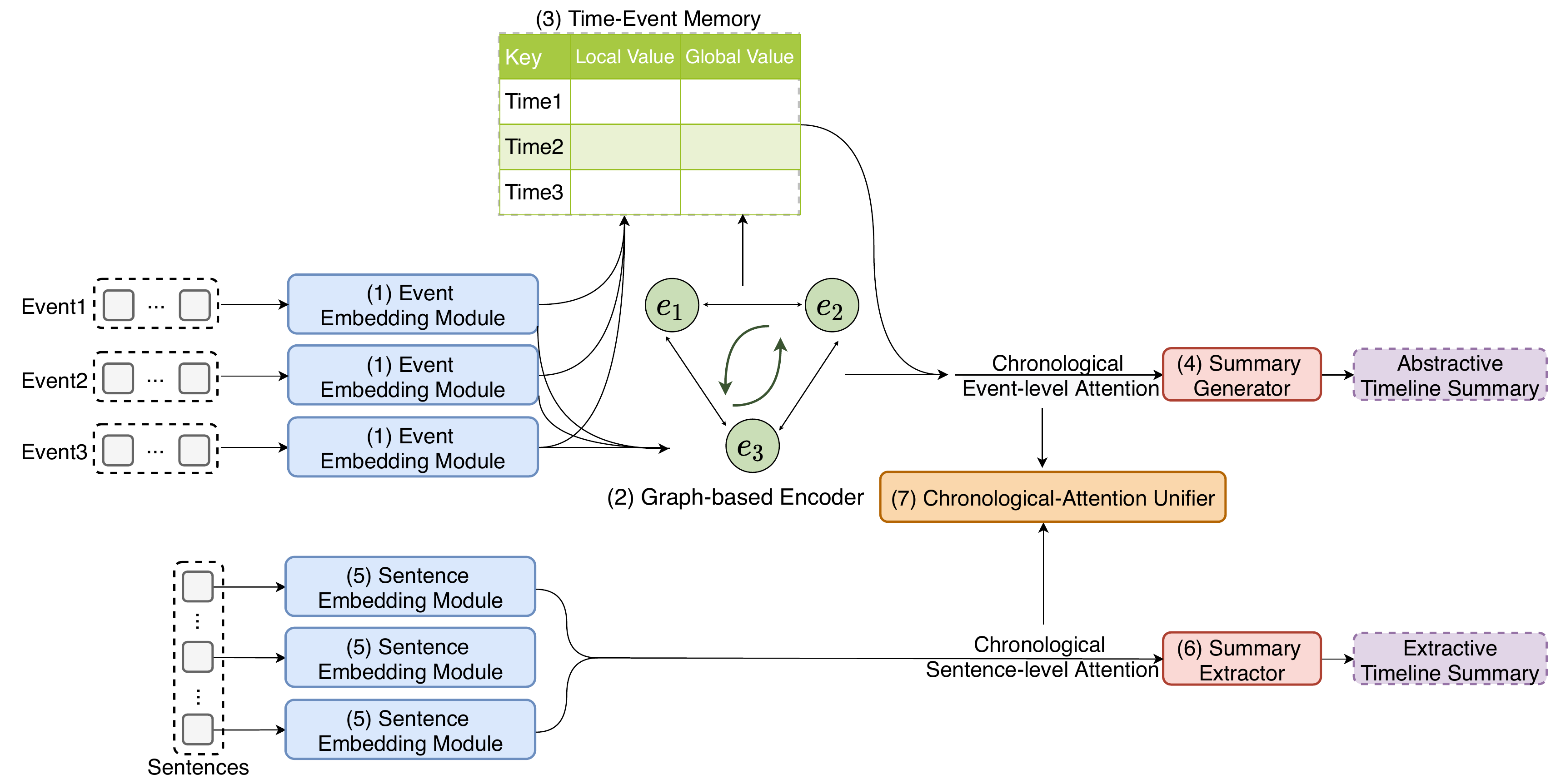}
		\caption{
			Overview of UTS. We divide our model into abstractive summarization part and extractive summarization part.
			Abstractive part includes: 
			(1) \textit{Event Embedding Module},
			(2) \textit{Graph-based encoder},
			(3) \textit{Time-Event Memory}, and
			(4) \textit{Summary Generator}.
			Extractive part includes:
			(5) \textit{Sentence Embedding Module} and (6) \textit{Sentence Extractor}.
			Additionally, there is a (7) \textit{Chronological Attention Unifier} that unifies the two tasks.
		}
		\label{fig:overview}
	\end{figure*}

	In this section, we introduce our\textit{ Unified Timeline Summarizer} (UTS) in detail. 
	The overview of UTS is shown in Figure~\ref{fig:overview} and can be split into two parts, one aims to generate an abstractive summary, and one targets selecting important sentences as a summary.
	
    Abstractive part includes:
	(1) Event Embedding Module (See \S~\ref{subsec:embedding}): To obtain the vector representations for each event, we employ a recurrent network with Selective Reading Units (SRU) to learn the local representations.
	(2) Graph-based Encoder (See \S~\ref{subsec:graph}): The representations learned in the last module do not incorporate interaction between events. Hence, we propose a graph-based encoder to learn the global representation of each event incorporating the information from other events and the relationship between them. 
	 (3) Time-Event Memory (See \S~\ref{subsec:memory}): 
	we propose a time-event memory, which stores the local and global event representation, with time position keys together forming a timeline.
	(4) Summary Generator (See \S~\ref{subsec:generator}): eventually, we use an RNN-based decoder to generate the summary under the guidance of event-level attention and word-level attention.
	
	Extractive part includes:
	(5) Sentence Embedding Module (See \S~\ref{subsec:sentence}): this module embeds the sentence to a vector representation in a similar way to the event embedding module. 
	(6) Sentence Extractor (See \S~\ref{subsec:extractor}): the sentence extractor selects the salient sentences as the summary following the sequential time order.
	
	Additionally, we propose (7) Chronological-Attention Unifier (See \S~\ref{subsec:unify}), to let the two parts complement each other by unifying the attention distributions of abstractive parts and extractive parts.
	Concretely, we propose a time-aware inconsistency loss to penalize the inconsistency between these two tasks.
	
	Although some encoder and decoder modules in MTS are similar to UTS, there are three significant differences in our UTS model compared with MTS:
	\begin{enumerate}
    \item MTS encodes each event independently, without considering the information interaction between events.
    While in UTS, we propose a graph encoder, which learns global representations for input events.
    \item We propose a unified timeline framework that can not only generate an abstractive summary, but also an extractive summary.
    That is, only UTS includes the extractive part.
    \item We propose to unify the abstractive and extractive parts together, where the two tasks can benefit each other.
\end{enumerate}
	
Specifically, we show the comparison between MTS and UTS in Table~\ref{tab:comp_model}.

\begin{table}[t]
    \centering
    \caption{Comparision between MTS and UTS.}
    \begin{tabular}{@{}lcc@{}}
    \toprule
    & MTS & UTS \\ 
    \midrule
   Event Embedding Module & SRU & SRU \\
    Graph-based Encoder & - & Transformer \\
    Time-Event Memory & Key-Value Memory & Key-Value Memory \\
    Summary Generator & Editing Gate & Editing Gate \\
    Sentence Embedding Module  & - & SRU \\
    Sentence Extractor & -& RNN \\
    Unifier & - & Inconsistency loss \\
    \bottomrule
    \end{tabular}
    \label{tab:comp_model}
    \end{table}
	
	\subsection{Event Embedding Module}
	\label{subsec:embedding}
	
	We first propose an event embedding module to obtain the word-level and event-level vector representations.
	To begin with, we use an embedding matrix $e$ to map a one-hot representation of each word in $x_i$ into a high-dimensional vector space.
	We denote $e(w^i_t)$ as the embedding representation of word $w^i_t$.
	We then employ a bi-directional recurrent neural network (Bi-RNN) to model the temporal interactions between words:
	\begin{align}
	\overleftarrow{h^i_t} &= \text{LSTM}^\text{enc}([e(w^i_t) ; p^i], \overleftarrow{h^i_{t-1}}),\\
	\overrightarrow{h^i_t} &= \text{LSTM}^\text{enc}([e(w^i_t) ; p^i],\overrightarrow{h^i_{t-1}} ),\\
	h^i_{t}&=\overrightarrow{h^i_{t}} +\overleftarrow{h^i_{t}},
	\end{align}
	\noindent where ``;'' denotes the concatenation between vectors, and $h^i_t$ denotes the hidden state of $t$-th word in Bi-RNN for event $x_i$.
	To capture the sequential information of events, we randomly initialize a \emph{time position encoding} vector $p^i$ of $i$-th event to be included in the Bi-RNN input.
	
	Apart from obtaining word representation $h^i_t$, we also need to gain event representation.
	Simply taking the final state of Bi-RNN $h^i_{T^i_w}$ as the representation of the whole event cannot fully capture the feature of the whole event. 
	Thus, we employ the selective reading module consisted of SRU proposed in ~\cite{chen2018iterative} to gain new event representation $a^i$:
	\begin{align}
	\label{SRU}
	s_t^i&=\text{SRU}(s_{t-1}^i,[h_t^i,h^i_{T^i_w}]),\\
	a^i&=s^i_{T_w^i},
	\end{align}
	where $s_t^i$ is the hidden state of $t$-th SRU cell in $i$-th event.
	At the high level, SRU is a modified version of GRU, which replaces the update gate in original GRU~\citep{cho2014learning} with a new gate taking each input $h_t^i$ and coarse event representation $h^i_{T_w}$ into consideration.
	We omit the details here due to limited space and readers can refer to \cite{chen2018iterative} for details.
	So far, we obtain the representation of $i$-th event $a^i$ and $t$-th word in $a^i$, \emph{i.e.}, $h^i_t$.

	\subsection{Graph-based Encoder}
	\label{subsec:graph}
	The event representation $a^i$ in the previous section is calculated independently, without considering the information flow between different events.
	However, the importance of each event and whether it should be included in the summary does not only depend on itself but also is related to other events.
	For example, in Table~\ref{tab:intro-case}, Jackson releases his first solo album might be an important event, but its importance is weakened by his ``Thriller'' album that breaks the racial barriers.
	Hence, we propose a graph-based encoder to learn the relationship between events and obtain a global event representation that incorporates such information.
	
	As shown in Figure~\ref{fig:overview},
	to embed relationship information, we set up the relation edges in our document modeling graph.
	The relation edge in our graph is firstly initialized by the event representation:
	\begin{align}
	r^{i,j}=\text{MLP}_a([a^i;a^j]),
	\end{align} 
	where MLP is a multi-layer perceptron.
	
	Next, during the relation-aware encoding process, we incorporate the relation edge $r^{i,j}$ into the final event representation by self attention operation:
	\begin{align}
	b^i=\text{RE}(a^i,a^*,r^{i,*}),
	\end{align}
	where $*$ denotes all indexes between $1$ and ${T_e}$.
	This module is based on Transformer. Thus, we first introduce Transformer:
	\begin{align}
	b^{i'}=\text{Transformer}(a^i,a^*).
	\end{align}
	Concretely, the first input is for query and the second input is for keys and values.
	Each output element, $b^{i'}$, is computed as weighted sum of a linearly transformed input values:
	\begin{align}
b^{i'} =  \sum_{j=1}^{T_e} \alpha^{i,j}_g  \left(a^jW^V\right).\label{equ:transformer-sum}
	\end{align}
	Each weight coefficient, $\alpha^{i,j}_g$, is computed using a softmax function:
	\begin{align}
	\alpha_g^{i,j}=\frac{\exp \left(\beta^{i,j}_g\right)}{\sum_{k=1}^{T_e} \exp  \left(\beta^{i,k}_g\right)}.
	\end{align}
	$\beta^{i,j}_g$ is computed using a compatibility function that compares two input elements:
	\begin{align}
	\beta^{i,j}_g=\frac{\left(a^{i} W^{Q}\right)\left(a^j W^{K}\right)^{T}}{\sqrt{d}}\label{eq:alpha},
	\end{align}
	where $d$ is the hidden dimension, and $W^Q, W^K, W^V \in \mathbb{R}^{d \times d}$ are parameter matrices.

	RE is similar to Transformer, with two changes in Equation~\ref{equ:transformer-sum} and  \ref{eq:alpha}.
	Specifically, we modify Equation~\ref{equ:transformer-sum} to propagate edge information to the sub-layer output:
	\begin{align}
	b^i =  \sum_{j=1}^{T_e} \alpha^{i,j}_g  \left(a^jW^V_r+r^{i,j}\right).
	\end{align}
	In this way, the representation of each event is more comprehensive, consisting of its relation dependency information with other events.
	In the meantime, when deciding the weight of each edge, \textit{i.e.,} $\beta^{i,j}_g$, we also incorporate relation edge information, since close relationships can have a great impact on edge weight.
	Concretely, Equation~\ref{eq:alpha} is changed to:
	\begin{align}
	\beta^{i,j}_g=\frac{\left(a^i W^{Q}_r\right)\left(a^j W^{K}_r+r^{i,j}\right)^{T}}{\sqrt{d}}.
	\end{align}

	The intuition for Transformer architecture is that each input is not isolated, and its representation depends on other inputs as well.
	In our augmented Transformer, \textit{i.e.,} graph-based encoder, the polished event representation $b^i$ follows the same idea and expands the dependency between input documents.
	$b^i$ does not only depend on its corresponding content but also depends on other inputs, as well as the relationships with others.

	\subsection{Time-Event Memory}
	\label{subsec:memory}
	
	As stated in the Introduction, in the timeline dataset, the generated summary should capture the time-series information to guide the chronological generation process.
	Hence, we propose a key-value memory module where keys together form a timeline, and this time series information is used to guide the generation process as shown in Figure~\ref{fig:decoder}.
	
	The key in this memory is the time position encoding $p^i$ introduced in \S~\ref{subsec:embedding}.
	We will use this key as time guidance to extract information from the value part in the memory, which will be introduced in detail in \S~\ref{subsec:generator}.
	The value part stores event information of local aspect in local value and global aspect in global value.
	Local value simply stores the event representation $a^i$, which means that only captures information from the current event.
	On the other hand, the global value is responsible for learning the event feature from a global perspective, not only based on itself but also its relationship with other events.
	Hence, it stores the graph-based encoder output, $b^i$.

	\subsection{Summary Generator}
	\label{subsec:generator}
	
\begin{figure}
		\centering
		\includegraphics[scale=0.5]{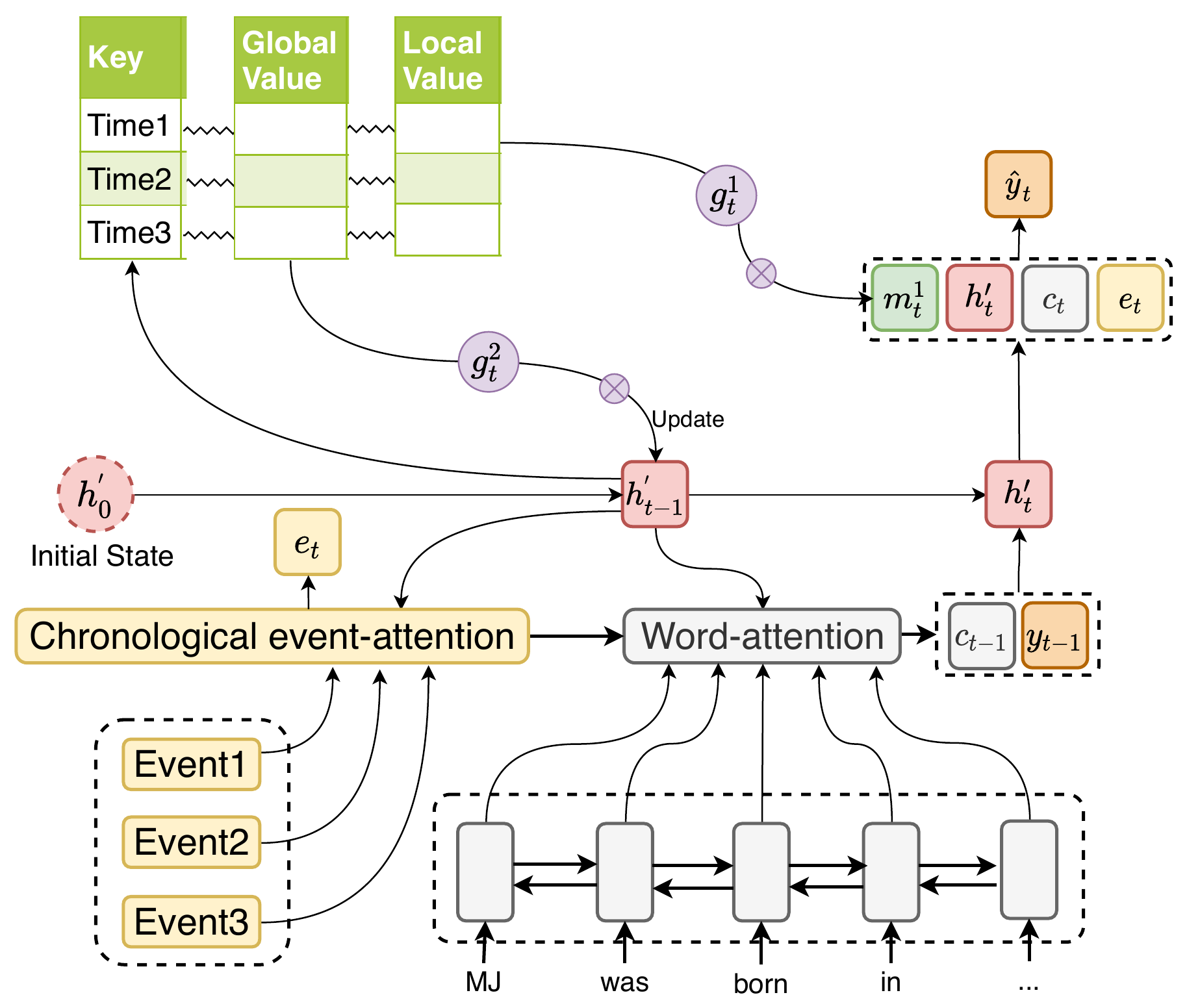}
		\caption{
			An overview of the summary generator in the abstractive part. The summary generator generates the next word based on word-level and event-level attention, as well as the key-value memory. 
		}
		\label{fig:decoder}
	\end{figure}

	To generate a consistent and informative summary, so as to avoid mixing information from different time stamps due to unawareness of correct timeline, we propose an RNN-based decoder that incorporates outputs of time-event memory module and event representation as illustrated in Figure~\ref{fig:decoder}.
	
	Following \cite{li2018survey}, we randomly initialize an LSTM cell taking the concatenation of all event representations as input, and use the output as decoder initial state:
	\begin{align}
	h'_0 &= \text{LSTM}^{\text{ini}} \left( h_c, [a^1; ...;  a^{T_{e}}] \right),  \label{eq:init-state}
	\end{align}
	where $h_c$ is a random variable.
	
	\subsubsection{Word-level attention}
	\label{subsec:guidance}
	Next, following traditional attention mechanism in ~\cite{Bahdanau2015Neural}, we summarize the input document into context vector $c'_{t-1}$ dynamically, and the $t$-th decoding step is calculated as:
	\begin{align}
	h'_t &= \text{LSTM}^{\text{dec}} \left(h'_{t-1}, [c'_{t-1} ; e(y_{t-1})]\right), \label{eq:dec-step}
	\end{align}
	where $h'_t$ is the hidden state of $t$-th decoding step.
	Context vector $c'_{t-1}$ is calculated as:
	\begin{align}
	\alpha^{t'}_{i,j} &= W_a^\intercal \tanh \left( W_b h'_{t-1} + W_h h^i_j \right), \\ 
	\alpha^{t}_{ i,j} &= \exp \left( \alpha^{t'}_{i,j} \right) /  \sum^{T_e}_{k=1}\left( \sum^{T^i_w}_{j=1}\exp \left(\alpha^{t'}_{ k,j} \right)\right), \label{equ:attention-sm}\\
	c'_{t-1} &= \sum_{i=1}^{T_{e}}  \left(\sum_{j=1}^{T^i_w} \alpha^{t}_{ i,j} h^i_j\right) \label{eq:context},
	\end{align}
	where we first use the decoder state $h'_{t-1}$ to attend to each states $h^i_j$ which results in the attention distribution $\alpha^t_{i,j}$, shown in  Equation~\ref{equ:attention-sm}.
	$h^i_j$ denotes the representation of $j$-th word in event $x_i$.
	Then we use the attention distribution $\alpha^t_{i,j}$ to obtain the weighted sum of document states as the context vector $c_{t-1}'$.

	Context vector $c'_{t-1}$ here only takes the word-level attention into consideration without considering event-level information.
	However, in timeline summarization, it is important for the model to be aware of which event it is currently describing, or it may confuse information from different events and result in an unfaithful summary.
	Hence, we introduce an event-level attention $\beta$ similar to the calculation of word-level attention and use it to adjust word-level attention:
	\begin{align}
	\beta^{t'}_{i} &= W_c^\intercal \tanh \left( W_d h'_{t-1} + W_e a^i \right), \\ 
	\beta^{t}_{i} &= \exp \left( \beta^{t'}_{i} \right) /  \sum^{T_e}_{j=1} \exp \left(\beta^{t'}_{j} \right),  \label{equ:beta}\\ 
	\gamma^t_{i, j} &= \alpha^{t}_{i,j}  \beta^{t}_{i}.
	\end{align}
	The new context vector $c_{t}$ (replacing $c_{t}'$ in Equation~\ref{eq:dec-step}) is now calculated as:
	\begin{align}
	c_{t} &= \sum_{i=1}^{T_{e}}  \left(\sum_{j=1}^{T^i_w} 	\gamma^t_{i, j}  h^i_j\right).
	\end{align}
	
	\subsubsection{Event-level attention}
	\label{subsec:chronological}
	Apart from using event-level attention to directly guide word-level attention, we also use it to obtain the weighted sum of event representation to be concatenated in the projection layer in Equation~\ref{equ:out-proj}:
	\begin{align}
	e_t= \sum_{i=1}^{T_e}{\beta^t_{i}a^i}.
	\end{align}

	\subsubsection{Memory guidance}
	
	So far, we have finished the calculation of the context vectors.
	Next, we introduce how to incorporate the guidance from memory.
	We first use hidden state $h'_t$ to attend to each key in memory.
	As stated in \S~\ref{subsec:memory}, keys, \emph{i.e.}, time position embeddings, conform the timeline that represents the time series information.
	Thus, we let the model take advantage of this sequential information, and calculate the relevance between position encoding and current state as time-attention $\pi(p^i, h'_t)$:
	\begin{align} \label{equ:attr-key-score}
	\pi(p^i, h'_t) = \exp(h'_t W_e p^i)/ \sum^{T_e}_{j=1} \exp(h'_t  W_e p^j).
	\end{align}
	Time-attention is then used to gain the weighted sum of local value $v_1$ and global value $v_2$ in the memory:
	\begin{align}\label{equ:read-kvmn-value}
	m^{1'}_t &=  \sum^{T_e}_{i=1} \pi(p^i,h'_t)v^i_1,\\
	m^{2'}_t &=  \sum^{T_e}_{i=1} \pi(p^i,h'_t)v^i_2.
	\end{align}
	$m^{1'}_t$ and $m^{2'}_t$ stores information from different level, thus should play different roles in generator.
	
	By a fusion gate, local value $m^{1'}_t$ is changed to $m^{1}_t$ and will be incorporated into the projection layer in Euqation~\ref{equ:out-proj}.
	\begin{align}
	g^{1}_t &= W_o([h'_{t};c_t;m^{1'}_t]), \\
	m^{1}_t &= g^{1}_t\cdot  m^{1'}_t.
	\end{align}
	We place the local value in the projection layer since $m^{1}_t$ stores the detailed information rather than the global feature in the input, thus should play an important part when generating each word.
	
	As for the global value $m^{2'}_t$, it stores the global feature of the event in a different position, thus should influence the whole generation process.
	Concretely, information from $m^{2'}_t$ is fusioned into the decoding state $h_{t}'$ by a gate:
	\begin{align}
	g^{2}_t &= W_n([h_{t}';c_t;m^{2'}_t]), \\
	h_{t}' &= g^{2}_t\cdot  h_{t}' + (1-g^{2}_t) \cdot m^{2'}_t.\label{eq:fusionhidden}
	\end{align}

	Finally, an output projection layer is applied to get the final generating distribution $P_{v}$ over vocabulary:
	\begin{align}
	P_{v} = \text{softmax} \left( W_v [ m^1_t;h'_t;c_t;e_t]  + b_v \right).
	\label{equ:out-proj}
	\end{align}
	We concatenate the output of decoder LSTM $h'_t$, the word context vector $c_{t}$, the event context vector $e_t$, and memory vector $m^1_{t}$ as the input of the output projection layer.
	
	In order to handle the out-of-vocabulary (OOV) problem, we equip the pointer network~\citep{Gu2016IncorporatingCM,getto17} with our decoder, which enables the decoder capable of copying words from the source text.
	The design of the pointer network is the same as the model used in~\cite{getto17}, thus we omit this procedure due to limited space.
	
	Our objective function in the abstractive part is the negative log likelihood of the target word $y_t$, shown in Equation~\ref{eq:loss-generator}:
	\begin{align}
	\mathcal{L}^{\text{abs}} &= -  \sum^{T_y}_{t=1} \log P_{v}(y_t). \label{eq:loss-generator}
	\end{align}
	The gradient descent method is employed to update the parameters in the abstractive part to minimize this loss function.

	\subsection{Sentence Embedding Module}
	\label{subsec:sentence}
	So far, we introduce the abstractive timeline summarization part.
    Next, we will introduce the extractive summarization part in UTS, and how to unify these two tasks.
	
	Our sentence embedding module takes 
	inspiration from \citep{chen2018iterative}, where the embedding module also takes the form of a hierarchical structure and consists of an iterative polishing process to better encoder the input document.
	Concretely, we employ a new Bi-RNN to process each sentence and obtain the representation $\hat{h}_t^i$, denoting the $t$-th word in $i$-th sentence.
	We use the last hidden state to represent the overall sentence representation, denoted as $\hat{h}^i_{T_{w}}$. 
	The document representation is initialized as the average of all sentence representations:
	\begin{align}
	D_{1}=\tanh \left(W \frac{1}{T_{ys}} \sum_{i=1}^{T_{ys}}\left[\hat{h}^i_{T_w}\right]+b\right).
	\end{align}
	Next, to model the sequential relationship between sentences and obtain a more comprehensive sentence representation, we iteratively polish the sentence and document representations.
	For brevity, we take the first iteration as an example to illustrate the process.
	Concretely, there is an RNN based on SRU (introduced in Equation~\ref{SRU}) in the iteration:
	\begin{align}
	\hat{a}_1^i&=\text{SRU}(\hat{a}_0^{i-1},[\hat{h}^{i-1}_{T_w},D_1]),\\
	D_2&=\text{GRU}_{\text{iter}}(\hat{a}_1^{T_{ys}},D_1),
	\end{align}
	where $\hat{a}^i_1$ is the hidden state of $i$-th SRU cell in the first iteration.
	In this way, we iteratively polish the sentence and document representation.
	We use $I$ to denote the iteration number, thus the final representation for $i$-th sentence is $\hat{a}_I^{i}$.

	\begin{figure}
		\centering
		\includegraphics[scale=0.65]{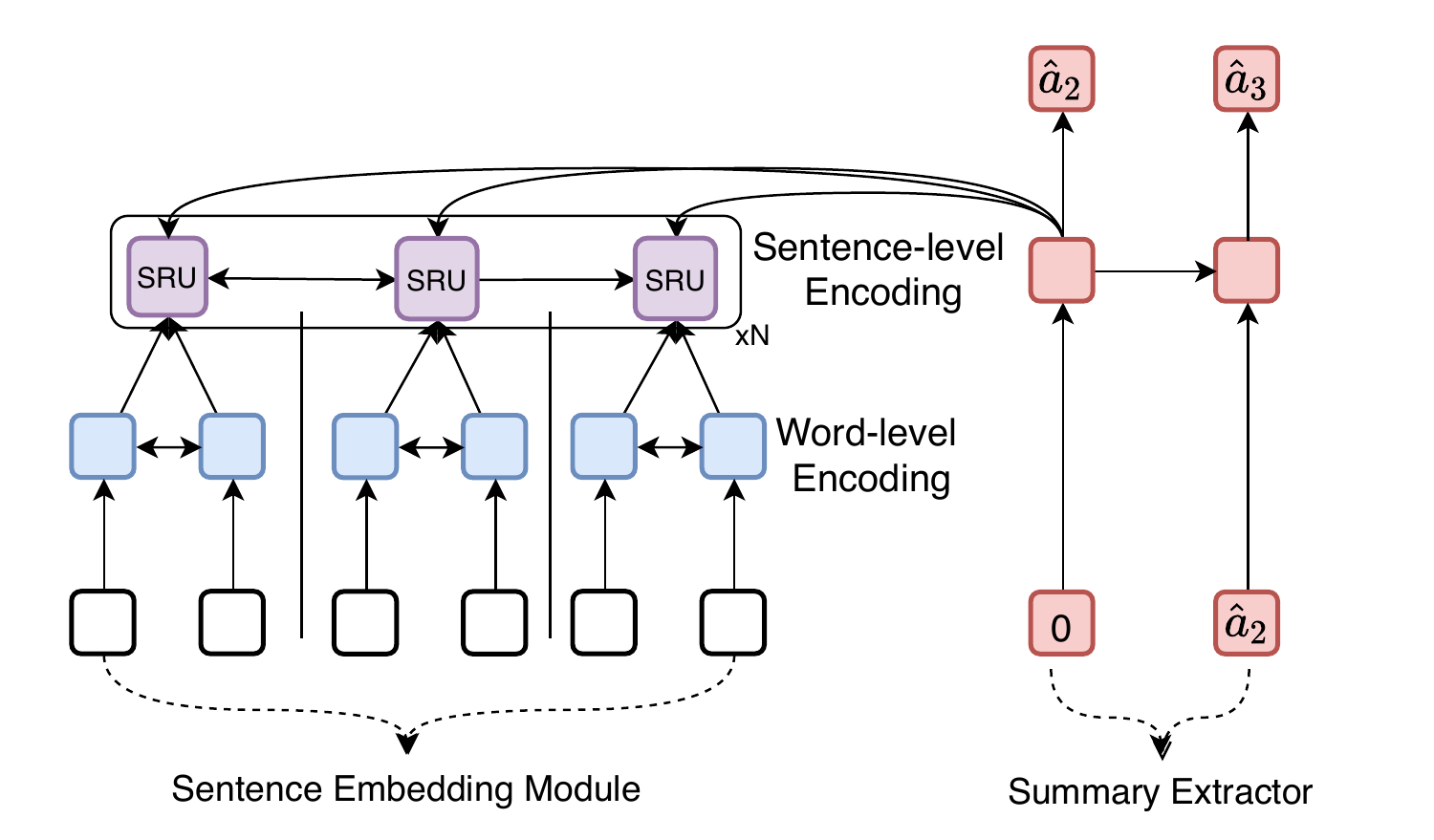}
		\caption{
			An overview of the sentence extractor in the extractive part. 
			In each decoding step, a sentence is to be included in the summary in sequence.
		}
		\label{fig:decoderb}
	\end{figure}

	\subsection{Sentence Extractor}
	\label{subsec:extractor}
	Different from previous work that builds a classifier to assign importance score to each sentence, we use an RNN consisting of LSTM cells to select sentences, wherein each step a sentence is selected as illustrated in Figure~\ref{fig:decoderb}.
Following traditional attention mechanism in \cite{Bahdanau2015Neural}, we summarize the input document sentences into context vector $c_{\text{ext}}$ dynamically, and the $t$-th decoding step is calculated as:
	\begin{align}
	\tilde{h}_{t+1}&=\text{LSTM}^{\text{ext}}(\tilde{h}_{t},[c_{t}^{\text{ext}};\hat{a}^{ot}_{I}]),\\
	c_{t}^{\text{ext}}&=\sum^{T_{ys}}_{i=1}\hat{\beta}_t^{i}\hat{a}^i_I,\\
	ot&=\text{argmax}(\text{MLP}(\tilde{h}_{t})),
	\end{align}
	where $ot$ is the index of the selected in $t$ step, and $\hat{a}^{ot}_{I}$ is the hidden state of the previously selected sentence.
	$\hat{\beta}_t^i$ is the attention weight on $i$-th sentence in $t$-th step, and is computed in a similar way to \S\ref{subsec:guidance}.
	Thus, the details are omitted here due to limited space.

	In this way, our extracted summary is generated in sequence, so as to better capture the sequential information in the input:
	\begin{align}
	\mathcal{L}^{\text{ext}} &= - \sum_{i=1}^{T_{ys}} \log P_s\left(l_{i} \right).
	\end{align}

	\subsection{Chronological-Attention Unifier}
	\label{subsec:unify}
	In both abstractive and extractive timeline summarization tasks, the attention on the input document should both follow the time sequential order.
	Hence, it is intuitive to encourage these two levels of attention to be mostly consistent with each other during training as an intrinsic learning target for free (\ie without additional human annotation).

\begin{figure}
	\centering
	\includegraphics[width=0.99\linewidth]{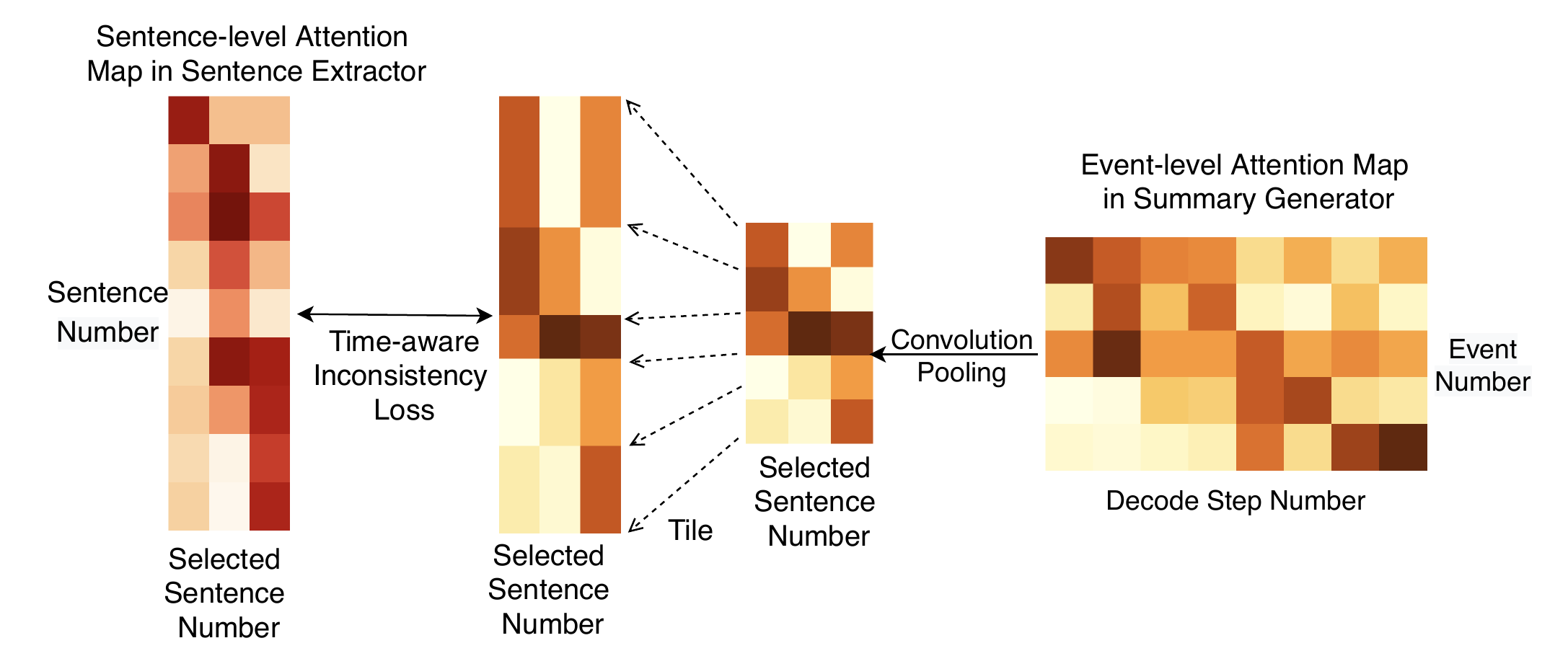}
	\caption{The illustration of the Chronological-Attention Unifier.
	After a convolution and a tile operation, the event-level attention in the summary generation is compared with the sentence-level attention in sentence extractor, where a novel inconsistency loss function is introduced to penalize the inconsistency between these two levels of attentions.}
	\label{fig:timeline-extract}
\end{figure}

    In \S\ref{subsec:guidance}, we propose the event-level attention $\beta$ in abstractive part, while in \S\ref{subsec:extractor}, the extractor pays sentence-level attention $\hat{\beta}$ on the input.
	The event-level attention evolves each time a new word is predicted in the summary generator, while the sentence-level attention evolves when a new sentence is selected.
    Hence, we first use a convolutional neural network (CNN) to extract the evolving event attention feature from the generator.
	Concretely, as shown in Figure~\ref{fig:timeline-extract}, a convolution along the decode step number axis is conducted on the event-level attention map, and the new attention matrix with the sentence-numbered axis is obtained. 
	Then, for each sentence-select step, we duplicate the event-level attention $\beta^i_t$ multiple times, where the duplicate number is the sentence number in the $i$-th event.
	
	Finally, we would like the event-level attention to be high when the sentence-level attention is high. 
	Hence, we design the following time-aware inconsistency loss:
	\begin{align}
	\mathcal{L}^{\text{inc}}=-\frac{1}{T_y} \sum_{t=1}^{T_{ys}} \log \left(\frac{1}{|\mathcal{K}|} \sum_{t \in \mathcal{K}} \hat{\beta}_t \times \beta_t\right),
	\end{align}
	where $\mathcal{K}$ is the set of top K attended sentences.
	This implicitly encourages the distribution of the sentence-level attentions to be sharp and event-level attention to be high. 
	To avoid the degenerated solution for the distribution of sentence attention to be one-hot and event attention to be high, we include the original loss functions for training the extractor ($\mathcal{L}^{\text{ext}}$ in \S\ref{subsec:extractor}) and abstracter ($\mathcal{L}^{\text{abs}}$ in \S\ref{subsec:generator}). 
	Note that this module is the only part that the extractor is interacting with the abstracter.
	 Our time-aware inconsistency loss facilitates our end-to-end trained unified model to be mutually beneficial to both the extractor and abstracter.

	\newcommand{\cbkgrnd}{\cellcolor{blue!15}}
	\section{Experimental Setup}
	\label{sec:setup}
	\subsection{Research Questions}
	We list seven research questions that guide the experiments: 
	\noindent \textbf{RQ1} (See \S~\ref{subsec:Overall}): What is the overall performance of UTS?
	Does it outperform other baselines on multilingual datasets? 
	\noindent \textbf{RQ2} (See \S~\ref{subsec:tl17}):
	What is the performance of our model on out-of-domain classic timeline summarization dataset?
	\noindent \textbf{RQ3} (See \S~\ref{subsec:ablation}):  What is the effect of each module in UTS?
	Does our multi-task framework help better summarization performance?
	\noindent \textbf{RQ4} (See \S~\ref{subsec:Fidelity}): Is the time position embedding useful so that the summary generator can attend to correct information in the time-event memory?
	\noindent \textbf{RQ5} (See \S~\ref{sec:two}): Can event-level attention correctly guide word-level attention in decoding process in the abstractive part?
	\noindent \textbf{RQ6} (See \S~\ref{sec:unify}): Are the chronological attentions successfuly unified in the abstractive and extractive summarization tasks?
	\noindent \textbf{RQ7} (See \S~\ref{robustness}): What is the influence of the parameter settings?
	
	\subsection{Dataset}
	To our best knowledge, there are no large-scale summarization datasets for timeline.
	Hence, in our previous work~\citep{Chen2019Learning}, we collect a large-scale timeline dataset from the world's largest Chinese encyclopedia\footnote{\url{https://baike.baidu.com/}}.
	The character subsection of this website consists of celebrities at all times and in all countries or lands.
	On each website page, there is a timeline summary for each character, and in the character experience section of this page, each event is set as a paragraph with explanation and details, which is selected as an input article.
	In the previous timeline works \cite{yan2011timeline}, they did not pre-select important sentences from the event news articles as a way to test the summarization ability of the proposed model.
	Hence, in our work, we did not preprocess the event paragraph as well, since these event paragraphs are similar to news articles in content and in style.
	We filter out irrelevant content such as cited sources and figures.
	We did not specifically extract time information from the input, because our model learns the time information in an implicit way, instead of particularly encoding it.
	In total, the training dataset amounts to 169,423 samples with 5,000 evaluation and 5,000 test samples.
	On average, there are 353.79 words and 61.19 words in the article and summary respectively.

	Furthermore, in this work, we first augment the previous dataset with event timeline summarization cases.
	On the Chinese encyclopedia, there is also a social event subsection that includes the developments of related events over time.
	Concretely, in the development history section of each page, there are event paragraphs that describe the development of the corresponding event, and there is also a corresponding timeline summary for these events.
	After the same cleaning operation, we have 83,188 training cases, 3,000 validation, and 3,000 test samples.
	On average, there are 495.19 words and 141.62 words in the article and summary respectively.

	Note that the above two datasets are both in the Chinese language. 
	To test the performance of our model on multi-lingual datasets, we collect an English timeline summarization dataset from Wikipedia websites.
	Since there are no character or event subsections in Wikipedia, we filter timeline pages by checking if there are multiple timestamps in the summary and document on each website.
	Other preprocesses are similar to the Chinese encyclopedia.
	A human evaluation on 200 sampled cases from the collected corpus shows that 196 cases are timeline document-summary pairs, 145 of which are about characters, and 51 are about events.
	
	Since we have large-scale English summarization datasets, we can test the generalization ability of our model on classic timeline summarization datasets, which are small-scale.
	Concretely, we report the performance of UTS on out-of-domain dataset Timeline 17 (TL17) \cite{Tran2013LeveragingLT}.
	TL17 contains human-written timelines about topics such as civil wars or the British Petroleum oil disaster, collected from major news outlets. 
	Each topic also has a set of related news articles scraped from the web.

	\begin{table}
		\begin{tabular}{ l | l | c c | c c | r r}
			\midrule
			\multirow{2}{*}{Datasets} & \multirow{2}{*}{\# docs (train/val/test)} & \multicolumn{2}{c|}{avg. document length} & \multicolumn{2}{c|}{avg. summary length} & \multicolumn{2}{c}{vocabulary size}\\
			& & words & sentences & words & sentences & document & summary\\ \midrule 
			TL17 & 4,650 & 1,252.33 & 63.76 & 43.66 & 2.73 & 102,099 & 6,725\\
			Celebrity TS & 169,423/5,000/5,000 & 353.79 & 12.76 & 61.19 & 3.97 & 444,725 & 191,334 \\
			Event TS & 83,188/3,000/3,000 & 495.19 & 18.73 & 141.62 & 6.05 & 1,083,249 & 368,619 \\ 
			Wiki TS &140,000/5,000/5,000& 606.65 & 27.79 &79.19 &7.89 &1,029,617 & 438,011\\ 
			\midrule
		\end{tabular}
		\caption{Comparison of summarization datasets with respect to overall corpus size, size of training, validation, and test set, average document (source) and summary (target) length (in terms of words and sentences), and vocabulary size on both on source and target.
		TS denotes Timeline Summarization.
		} 
		\label{table:bbc-size-comparison} 
	\end{table}
	
	The statistics of the four datasets are listed in Table~\ref{table:bbc-size-comparison}.
	We also give timeline statistic information in Table~\ref{tab:timeline}.
	It can be seen that compared with TL17 dataset, our three datasets are significantly larger.
	This again demonstrates the necessity of our datasets, which are large enough to train a neural-based model.
	In terms of timeline-related attributes, the summaries in our datasets have more date stamps in each sentence on average, which requires the summarization model to be more time-aware.
	The average date number in the document input is smaller in our datasets, this is because that our input document is shorter than TL17.
	However, the average sentences/dates ratio of our datasets is comparable to TL17, proving the time attribute of our datasets.

		\begin{table}
		\begin{tabular}{ l |  c c | c c | r r}
			\midrule
			\multirow{2}{*}{Datasets} &  \multicolumn{2}{c|}{Document} & \multicolumn{2}{c|}{Summary length} & \multicolumn{2}{c}{Compression}\\
			& Avg dates & Avg sents/dates &  Avg dates & Avg sents/dates& Sent & Date\\ \midrule 
			TL17  & 77.55 & 1.22 & 1.61 & 1.14 & 34.70 & 32.46 \\
			Celebrity TS  & 11.03 & 1.18 & 3.15 & 1.10 & 3.76 & 3.50\\
			Event TS  & 13.09 & 1.54 & 5.55 & 1.19 & 3.03 & 2.36 \\ 
			Wiki TS & 6.81 & 3.99 & 2.49 & 3.17 & 3.44 & 2.74 \\ 
			\midrule
		\end{tabular}
		\caption{Timeline-specific statistic attributes of our datasets and TL17 dataset.
		} 
		\label{tab:timeline} 
	\end{table}
	
	\subsection{Comparison Methods}
	
	We first conduct an ablation study to prove the effectiveness of each module in UTS.
	Then, to evaluate the performance of our proposed dataset and model, we compare it with the following baselines:
	
	Abstractive baselines:
	
	\noindent (1) \textbf{Pointer-Gen} \citep{getto17} is an RNN based model with an attention mechanism and allows the system to copy words from the source text via pointing for abstractive summarization.
	
	\noindent	(2) \textbf{FTSum} leverages open information extraction and dependency parse technologies to extract actual fact descriptions from the source text \citep{cao2018faithful}.
	Since there is no open information extraction tool in Chinese, we use POS tagging to extract entities and verbs to replace them.
	
	\noindent	(3) \textbf{Unified} is a unified model combining the strength of extractive and abstractive summarization proposed in \cite{unified18}, where sentence-level attention is used to modulate the word-level attention such that words in less attended sentences are less likely to be generated.
	
	\noindent	(4) \textbf{GPG} is a model proposed by \citet{shen2019improving} which generates summaries by ``editing'' pointed tokens instead of hard copying.
	The editing is performed by transforming the pointed word vector into a target space with a learned relation embedding.
	
	\noindent	(5) \textbf{SAGCopy} is an augmented Transformer with a self-attention guided copy mechanism, which was proposed by \citet{xu2020self}.
	Specifically, they first identify the importance of each source word based on the degree centrality with a directed graph built by the self-attention layer in the Transformer. 
	They then use the centrality of each source word to guide the copy process explicitly.
	
	\noindent	(6) \textbf{MTS} is the first abstractive timeline summarization framework proposed in our previous work~\cite{Chen2019Learning}. This method achieves state-of-the-art performance on the celebrity timeline summarization dataset.

	Extractive baselines:
	
	\noindent 	(1) \textbf{Lead3} is an extractive baseline that concatenates the first-3 sentences of each source document as a summary.
	
	\noindent	(2) \textbf{TextRank} ~\citep{Mihalcea2004TextRankBO} is an unsupervised algorithm while sentence importance scores are computed based on eigenvector centrality within weighted-graphs for extractive sentence summarization.
	
	\noindent	(3) \textbf{ITS} One of state-of-the-art extractive summarization models proposed in \cite{chen2018iterative}.
	ITS iteratively polishes the document representation on many passes through the document, so as to extract better summaries.
	
	For testing our models on out-of-domain dataset, we compare with a number of traditional timeline summarization baselines:
	
	\noindent	(1) \textbf{Chieu} \cite{chieu2004query} is an unsupervised baseline based on direct summarization. 
	
	\noindent	(2) \textbf{Martschat} \cite{martschat2018temporally} greedily selects a combination of sentences from the entire collection, which maximizes submodular functions for content coverage, textual and temporal diversity and a high count of date references.
	
	\noindent	(3) \textbf{Tran}\cite{binh2013predicting} is an original date-wise timeline summarization approach, using regression for both date selection and summarization, and using all sentences of a date as candidate sentences.
	
	\noindent (4) \textbf{Pubcount} \cite{Ghalandari2020ExaminingTS} is a simple date-wise baseline that uses the publication count to rank dates, and all sentences published on a date for candidate selection. 
	
	\noindent	(5) \textbf{Datawise} \cite{Ghalandari2020ExaminingTS} uses supervised date selection, PM-MEAN for candidate selection and CENTROID-OPT for summarization.
	
	\noindent	(6) \textbf{Clust} \cite{Ghalandari2020ExaminingTS} uses DATEMENTIONCOUNT to rank clusters, and CENTROID-OPT for summarization.
	
	The performance of these baselines are consistent with the result from \citep{Ghalandari2020ExaminingTS}.
	
	\subsection{Evaluation Metrics}
	
	For evaluation metrics, we adopt ROUGE F1 score in ~\cite{lin2004rouge} which is widely applied for summarization evaluation~\citep{Sun2018AUM,chen2018iterative}. 
	The ROUGE metrics compare the generated summary with the reference summary by computing overlapping lexical units, including ROUGE-1 (unigram), ROUGE-2 (bi-gram), and ROUGE-L (longest common subsequence).
	
	For the out-of-domain test dataset, we follow \cite{Ghalandari2020ExaminingTS}, and use the specific timeline evaluation metric, \ie Alignment-based ROUGE F1-score, and Date F1-score.
	Alignment-based ROUGE F1-score compares the textual overlap between a system and a ground-truth timeline, while also considering the assignments of dates to texts.
	Date F1-score compares only the dates of a system and a ground-truth timeline.
	
	\cite{E17-2007} notes that only using the ROUGE metric to evaluate summarization quality can be misleading. 
	Therefore, we also evaluate our model by human evaluation.
	Three highly educated participants are asked to score 100 randomly sampled summaries generated by \texttt{GPG}, \texttt{SAGCopy}, \texttt{MTS}, and UTS-\textit{abs}. 
	Statistical significance of observed differences between the performance of two runs are tested using a two-tailed paired t-test and is denoted using \dubbelop (or \dubbelneer) for strong significance for $\alpha = 0.01$.

	\subsection{Implementation Details}
	
	We implement our experiments in TensorFlow~\citep{abadi2016tensorflow} on NVIDIA GTX 1080 Ti GPU. 
	For all experiments, our model has 256-dimensional hidden states and 128-dimensional word embeddings. 
	Following \citet{getto17}, we do not pretrain the word embeddings, instead, they are learned from scratch during training.
	We use a vocabulary of 50k words for both source and target.
	For time-event memory, the dimension of the key, global value, and local value are 128, 512, and 256 respectively.
	We initialize all of the parameters randomly using a uniform distribution in [-0.02, 0.02].
	The batch size is set to 16, and the event number is set to 8.
	For the abstractive part, during training and at test time we truncate the article to 400 tokens and limit the length of the summary to 70 tokens.
	For the extractive part, we used a greedy algorithm similar to \cite{nallapati2017summarunner} to obtain an oracle summary for each document to train extractive models. 
	The algorithm generates an oracle consisting of multiple sentences which maximize the ROUGE-2 score against the gold summary.
	We limit the input sentence number to 24, the length of each input sentence to 20, and the number of selected sentences to 4.
	For the chronological-attention unifier, we set $K$ to 3 for computing $\mathcal{L}^{\text{inc}}$.
	We use Adagrad optimizer~\citep{Duchi2010AdaptiveSM} as our optimizing algorithm and the learning rate is 0.15.
	(This was found to work best of Stochastic Gradient Descent, Adadelta, Momentum, Adam, and RMSProp).
	 We use gradient clipping with a maximum gradient norm of 2, but do not use any form of regularization.
	 We use loss on the validation set to implement early stopping.
	In decoding, we employ a beam search with beam size 4 to generate a more fluent summary sentence.
	When testing our model on the out-of-domain Timeline 17 dataset, for each example with S source input documents, we take the first 400/S tokens from each source document.
	
	For the training efficiency, it takes about 9.7 hours to train an epoch, and our model reaches the best performance after only 3 epochs.
	While for baseline Pointer-Gen, it takes 7 hours to train an epoch, but it reaches the best performance after 7 epochs.
	In particular, our model makes much quicker progress in the early phases of training.
	This demonstrates the effectiveness of our unified model. 
	In terms of testing, it takes 1.06 hours to generate summaries for all the cases in the test dataset.
	We selected the top-3 checkpoints based on the evaluation loss on the validation set, and report the averaged results on the test set.

	\section{Experimental Results}
	\label{sec:results}
	
	\subsection{Overall Performance}
	\label{subsec:Overall}
	\newcommand{\phantomtriangle}{\phantom{\dubbelop}}

	For research question \textbf{RQ1}, we examine the performance of our model and baselines in terms of ROUGE as shown in table~\ref{tab:comp_rouge_baselines}.
	Firstly, on the celebrity timeline dataset, abstractive models outperform extractive models by a substantial margin on our datasets.
	We attribute this result to the observation that the gold summary of this dataset tends to use new expressions to summarize the original input documents.
	This demonstrates the necessity of abstractive timeline summarization approaches.
	Secondly, we compare our previous model \texttt{MTS} with recently-proposed baselines including \texttt{SAGCopy} and \texttt{GPG}.
	These two baselines obtain lower ROUGE scores on our datasets than \texttt{MTS}, which demonstrates the effectiveness of our previous model.
	Finally, based on \texttt{MTS}, our augmented model UTS-\textit{ext} and UTS-\textit{abs} achieves even better performance.
	
	Concretely, for the abstractive part, UTS-\textit{abs} outperforms \texttt{SAGCopy} by 7.56\%, 14.92\% and 7.61\%, and outperforms \texttt{MTS} by 4.47\%, 7.68\% and 3.95\% in terms of ROUGE-1, ROUGE-2 and ROUGE-L respectively on celebrity timeline dataset.
	For the extractive part, our extractive method achieves about 2.23\% points improvement on ROUGE-2 compared with \texttt{ITS} on the celebrity timeline dataset. 
	We attribute the improvement to two aspects: Firstly, the abstractive objective can promote the recognition of important sentences for the extractive model with the chronological attention unifier network. 
	Besides, while extractive gold label sequences are obtained by greedily optimizing ROUGE-2 on the gold-standard summary, gold labels may not be accurate. 
	Joint learning of two objectives may correct some biases for the extractive model due to the inaccurate labels.
	The above results prove the superiority of our model.
	Note that we mainly compare our model with \texttt{ITS}, because our extractive part is mostly based on \texttt{ITS}.
	Our framework can be applied to other extractive models, and theoretically, will bring benefits for both tasks.
	We leave it as future work.
	
		\begin{table}[t]
		\centering
		\begin{tabular}{l ccc ccc ccc}
			\toprule
			\multirow{3}{*}{Models} & \multicolumn{3}{c}{\textsf{Celebrity Timeline Dataset}} & \multicolumn{3}{c}{\textsf{Event Timeline Dataset}} & \multicolumn{3}{c}{\textsf{Wiki Timeline Dataset}}\\
			\cmidrule(lr){2-4}  \cmidrule(lr){5-7} 
			\cmidrule(lr){8-10} 
			& RG-1  & RG-2  & RG-L & RG-1  & RG-2  & RG-L
			& RG-1  & RG-2  & RG-L \\
			\midrule
			\multicolumn{4}{@{}l}{\emph{Sentence extraction methods}}\\
			Lead3      &   32.36  &17.96  & 30.99 &  21.47  &  9.26& 15.73 & 25.35 & 5.94 & 20.56  \\
			TextRank      &   32.27 &15.34 &  30.86 &  23.89  & 10.43 & 16.66  & 24.98 & 5.47 & 22.40 \\
			ITS      &   34.03  &18.20  &  31.24 &  27.94  &  14.28& 20.39 & 27.82 & 5.91 & 25.37 \\
			Unified-\textit{ext} &   34.18  &18.29  &  31.16 &  28.06 &  14.39& 20.47  &26.48 & 5.82 & 24.28  \\
			UTS-\textit{ext} & \textbf{34.81} & \textbf{22.26} &\textbf{32.03 }& \textbf{29.12} &  \textbf{16.01}& \textbf{23.06} & \textbf{29.00} & \textbf{6.64} & \textbf{25.81}\\
			\midrule
			\multicolumn{4}{@{}l}{\emph{Abstractive methods}}\\
			Pointer-Gen      &   36.61  &21.35  &   34.51 &  22.56 & 7.84& 21.00 & 23.12 & 5.07 & 19.65    \\
			FTSum  & 37.84 & 21.47 &35.37 & 23.41 & 6.95 & 21.66 & 24.08 & 5.80 & 20.05\\
			Unified-\textit{abs}  & 38.24 & 21.95 & 36.42& 23.58 &  7.93&21.95 & 24.34 & 5.84 & 20.37\\
			GPG & 38.43 & 21.59 &  36.38 & 22.38 & 7.77 &20.81 & 24.71 & 5.80 & 20.85 \\
			SAGCopy &38.64 &20.84& 36.41& 23.40
			 &7.95
			 &  21.72 & 26.00 & 5.84 & 22.01\\
			MTS & 39.78 & 22.24 & 37.69 & 23.89 & 8.38  &21.97 & 26.68 & 5.88 & 23.18 \\
			UTS-\textit{abs} & \textbf{41.56} & \textbf{23.95}& \textbf{39.18}& \textbf{25.30}
			 & \textbf{9.63}
			 & \textbf{23.28} & \textbf{27.71} & \textbf{5.92} & \textbf{24.62}\\
			\bottomrule
		\end{tabular}
			\caption{RQ1: ROUGE scores comparison between baselines. 
		Models and baselines in the top half are extractive, while those in the bottom half are abstractive.
		All our ROUGE scores have a 95\% confidence interval of at most $\pm$0.24 as reported by the official ROUGE script.}
		\label{tab:comp_rouge_baselines}
	\end{table}

		\begin{table}[t]
		\centering
		\begin{tabular}{@{}lccc@{}}
			\toprule
			& Fluency & Informativeness & Fidelity \\ 
			\midrule
			GPG & 2.59\phantom{0} & 2.53\phantom{0}&  2.39\phantom{0}\\
			\cbkgrnd SAGCopy & \cbkgrnd 2.64\phantom{0} &\cbkgrnd2.57 \phantom{0} &  \cbkgrnd 2.43\phantom{0}\\
			MTS& 2.71\phantom{0} & 2.58\phantom{0}& 2.61\phantom{0}\\
			UTS-\textit{abs} & \textbf{2.77}\dubbelop & \textbf{2.62}\dubbelop & \textbf{2.65}\dubbelop\\
			\bottomrule
		\end{tabular}
		\caption{RQ1: Human evaluation comparison with main baselines on celebrity timeline dataset.}
		\label{tab:comp_human_baslines}
	\end{table}

	Our human evaluation study assessed the overall quality of the summaries on the celebrity timeline dataset by asking three highly educated participants to rank them taking into account the following criteria: \textit{Fluency} (is the summary fluent and grammatical?), \textit{Informativeness} (does the summary convey important facts about the topic in question?), and \textit{Fidelity} (is the summary faithful to the input?).
	We pick \texttt{SAGCopy} and \texttt{GPG} as baselines since their performance is relatively high compared to other baselines.
	The rating score ranges from 1 to 3 and 3 is the best.
	The results are presented in Table~\ref{tab:comp_human_baslines}.
	We can see that our model performs much better than all baselines. 
	In the fluency indicator, our model achieves a high score of 2.77, which is higher than 2.59 of \texttt{GPG} and 2.64 of \texttt{SAGCopy}, indicating that our model can reduce the grammatical errors and improve the readability of the summary.
	In the informativeness indicator, our model is 0.05 better than \texttt{SAGCopy}. 
	It indicates that our model can effectively capture salient information.
	In the fidelity indicator, UTS-\textit{abs} outperforms all baselines by a large margin, which indicates the multi-granularity semantic information and joint learning with extractive summarization does help to avoid the unfaithful information of the generated summary.
	It is worth noticing that the infidelity problem is a serious problem existing in timeline summarization, and \texttt{MTS} and UTS-\textit{abs} greatly alleviates such problem.
	We also conduct the paired student t-test between our model and \texttt{SAGCopy} (the row with shaded background), and the result demonstrates the significance of the above results. 
	The kappa statistics is 0.46 and 0.49 respectively, which indicates moderate agreement between annotators\footnote{\cite{landis1977measurement} characterize kappa values $<$ 0 as no agreement, 0-0.20 as slight, 0.21-0.40 as fair, 0.41-0.60 as moderate, 0.61-0.80 as substantial, and 0.81-1 as almost perfect agreement.}.
	To prove the significance of these results, we also conduct the paired student t-test between our model and  \texttt{SAGCopy}. 
	We obtain a p-value of $3 \times 10^{-8}$, $8 \times 10^{-12}$, and $9 \times 10^{-11}$ for fluency, informativeness, and fidelity, respectively.

	\begin{CJK*}{UTF8}{gkai}
		\begin{table}[t]
			\centering
			\begin{tabular}{l|l}
				\toprule
				\multicolumn{2}{p{14.4cm}}{
					In \textcolor{red}{1981}, James Cameron directed his first film, ``Piranha II'', which was shot entirely in Italy. Cameron didn't get along well with an Italian speaking staff, and the producers didn't let him participate in the final editing of the film...
					In \textcolor{red}{1984}, Cameron released his first self-made and self-directed film ``Terminator'', which costs only 6.5 million dollars...
					In \textcolor{red}{1986}, James Cameron's second self-made work, ``Alien 2'', was published...
					In \textcolor{red}{1987}, ``Alien 2'' won seven Academy Award nominations...
					James Cameron won the best director award at the 14th Saturn Awards for this film.
					In \textcolor{red}{1989}, Cameron wrote and directed his third film, ``The Abyss''...
					In \textcolor{red}{1991}, his film ``Terminator 2'' made 200 million dollars at the box office in the United States, and he also won the 18th Saturn Awards for best director and best screenwriter for this film.
					In \textcolor{red}{1997}, Cameron directed the film ``Titanic'', which wins 1.84 billion at the box office, and starred Leonardo DiCaprio and Kate Winslet. 
					In \textcolor{red}{1998}, the film won 14 nominations and 11 awards at the 70th Academy Awards.
					In \textcolor{red}{2000}, he directed and supervised the TV series ``The dark angel of the last world'' with the theme of gene therapy.
				}  \\
				\hline
				reference                       & \multicolumn{1}{p{12.7cm}}{
					In \textcolor{red}{1981}, he directed the first film ``Piranha II''.
					In \textcolor{red}{1984}, he became famous for his science fiction film ``Terminator''. 
					In \textcolor{red}{1986}, he wrote and directed the film ``Alien 2''. 
					In \textcolor{red}{1991}, he won the best director award and best screenwriter award at the 18th Saturn awards for his film ``Terminator 2''. 
					In \textcolor{red}{1997}, his film ``Titanic'' won 1.84 billion dollars at the box office, breaking the global box office record; it won 11 awards including best picture at the 70th Academy Awards, and James Cameron won the best director award at the Oscars.}   \\ \hline
				GPG                         & \multicolumn{1}{p{12.7cm}}{
					In \textcolor{red}{1981}, James Cameron directed the film ``Piranha II''. 
					Cameron Cameron didn't get along well with an Italian speaking staff, and the producers didn't let him participate in the final editing of the film.
					In \textcolor{red}{1984}, he became famous for his science fiction film ``Terminator''. 
					In \textcolor{red}{1986}, he wrote and directed the film ``Alien 2''. 
					In \textcolor{red}{1986}, he wrote and directed the film ``Alien 2''. 
					In \textcolor{red}{1991}, he won the best director award and best screenwriter award at the 18th Saturn Awards for his film ``Terminator 2''.
					In \textcolor{red}{1997}, his film Titanic won 1.84 billion US dollars at the box office.
				}    \\ \hline
				SAGCopy                            & \multicolumn{1}{p{12.7cm}}{
					In \textcolor{red}{1981}, James Cameron directed his first work, ``Piranha II''.
					In \textcolor{red}{1984}, Cameron released his first film, ``Terminator''.
					In \textcolor{red}{1986}, James Cameron wrote and directed his second work, ``Alien 2''.
					In \textcolor{red}{1997}, James Cameron directed the film ``Titanic''.
					In \textcolor{red}{2000}, he directed and supervised the TV series ``The dark angel of the last world'' with the theme of gene therapy.
				} \\ \hline
				UTS-\textit{ext}                           & \multicolumn{1}{p{12.7cm}}{
					In \textcolor{red}{1981}, James Cameron directed his first film, ``Piranha II'', which was shot entirely in Italy.
					In \textcolor{red}{1984}, Cameron released his first self-made and self-directed film ``Terminator'', which cost only 6.5 million dollars.
					In \textcolor{red}{1986}, James Cameron's second self-made work, ``Alien 2'', was published.
					in \textcolor{red}{1997}, Cameron directed the film ``Titanic'', which wins 1.84 billion at the box office, and starred Leonardo DiCaprio and Kate Winslet.		
				} \\ \hline
				UTS-\textit{abs}                           & \multicolumn{1}{p{12.7cm}}{
					In \textcolor{red}{1981}, Cameron directed his first work, ``Piranha II''. 
					In \textcolor{red}{1984}, he released his first film ``Terminator''. 
					In \textcolor{red}{1986}, his second film, ``Alien 2'', was published. 
					In \textcolor{red}{1987}, ``Alien 2'' won the 14th Saturn Award for best director.
					In \textcolor{red}{1991}, his film ``Terminator 2'' made 200 million dollars in the United States.
					In \textcolor{red}{1997}, he directed the film ``Titanic''.
					In \textcolor{red}{1998}, the film won 14 Academy Awards nominations and 11 of them at the 70th Academy Awards.
				} \\ 
				\bottomrule
			\end{tabular}
			\caption{RQ1: Examples of the generated answers by UTS-\textit{abs}, UTS-\textit{ext} and baselines (translated version). }
			\label{tab:case-eng}
		\end{table}
	\end{CJK*}
	
		\begin{CJK*}{UTF8}{gkai}
		\begin{table}[t]
			\centering
			\begin{tabular}{l|l}
				\toprule
				\multicolumn{2}{p{14.5cm}}{
										\textcolor{red}{1981年}，詹姆斯·卡梅隆执导了第一部作品《食人鱼2》，影片完全在意大利拍摄。卡梅隆和一口意大利语的工作人员相处得并不愉快，而拍摄完毕后制片方不让他参与影片的最终剪辑.	\textcolor{red}{1984年}，卡梅隆推出了他第一部自编自导的影片《终结者》，这部影片的拍摄只花了650万美元...	\textcolor{red}{1986年}，詹姆斯·卡梅隆自编自导的第二部作品《异形2》问世...	\textcolor{red}{1987年}，《异形2》获得了七项奥斯卡奖提名...詹姆斯·卡梅隆凭借此片获得了第14届土星奖最佳导演奖...	\textcolor{red}{1989年}，卡梅隆自编自导了第三部电影《深渊》...	\textcolor{red}{1991年}，他执导的电影《终结者2》在美国上映后取得了2亿美元的票房，他也凭借该片获得了第18届土星奖最佳导演奖以及最佳编剧奖...	\textcolor{red}{1997年}，詹姆斯·卡梅隆执导了电影《泰坦尼克号》，该片获得18.4亿美元的票房，由莱昂纳多·迪卡普里奥、凯特·温斯莱特等主演.	\textcolor{red}{1998年}，在第70届奥斯卡金像奖上这部影片获得了14个奥斯卡奖的提名并获得了其中的11个奖项...	\textcolor{red}{2000年}，他执导并监制了以基因治疗为题材的电视剧《末世黑天使》...
							}  \\
				\hline
				reference                       & \multicolumn{1}{p{13cm}}{
										\textcolor{red}{1981年}，詹姆斯卡梅隆执导首部电影《食人鱼2》。
										\textcolor{red}{1984年}，因自编自导科幻电影《终结者》成名。	\textcolor{red}{1986年}，自编自导电影《异形2》。	\textcolor{red}{1991年}，凭借电影《终结者2》获得第18届土星奖最佳导演奖以及最佳编剧奖。
										\textcolor{red}{1997年}，他执导的电影《泰坦尼克号》取得了18.4亿美元的票房，打破全球影史票房纪录;该片在第70届奥斯卡金像奖上获得了包括最佳影片在内的11个奖项，詹姆斯·卡梅隆凭借该片获得了奥斯卡奖最佳导演奖。
					}   \\ \hline
				GPG                         & \multicolumn{1}{p{13cm}}{
										\textcolor{red}{1981年}，詹姆斯·卡梅隆执导了部作品《食人鱼2》，卡梅隆卡梅隆和一口意大利语的工作人员相处得并不愉快，而拍摄完毕后制片方不让他参与影片的最终剪辑.
										\textcolor{red}{1984年}，卡梅隆凭借科幻电影《终结者》出名。
										\textcolor{red}{1986年}，他自编自导了电影《异形2》。
							\textcolor{red}{1986年}，他自编自导了电影《异形2》。	
							\textcolor{red}{1991年}，他凭借《终结者2》获得了第18届土星奖最佳导演奖和最佳编剧奖。		\textcolor{red}{1997年}，他的电影《泰坦尼克号》在美国获得了18.4亿票房。
				}    \\ \hline
				SAGCopy                            & \multicolumn{1}{p{13cm}}{
									\textcolor{red}{1981年}，詹姆斯·卡梅隆执导了第一部作品《食人鱼2》。
									\textcolor{red}{1984年}，卡梅隆推出了他第一部自编自导的影片《终结者》。
									\textcolor{red}{1986年}，詹姆斯·卡梅隆自编自导的第二部作品《异形2》。
							\textcolor{red}{1997年}，詹姆斯·卡梅隆执导了电影《泰坦尼克号》。
								\textcolor{red}{2000年}，他执导并监制了以基因治疗为题材的电视剧《末世黑天使》。
				} \\ \hline
				UTS-\textit{ext}                           & \multicolumn{1}{p{13cm}}{
				\textcolor{red}{1981年}，詹姆斯·卡梅隆执导了第一部作品《食人鱼2》，影片完全在意大利拍摄。	\textcolor{red}{1984年}，卡梅隆推出了他第一部自编自导的影片《终结者》.这部影片的拍摄只花了650万美元。	\textcolor{red}{1986年}，詹姆斯·卡梅隆自编自导的第二部作品《异形2》问世。		\textcolor{red}{1997年}，詹姆斯·卡梅隆执导了电影《泰坦尼克号》，该片获得18.4亿美元的票房，由莱昂纳多·迪卡普里奥、凯特·温斯莱特等主演。
				} \\ \hline
				UTS-\textit{abs}                           & \multicolumn{1}{p{13cm}}{
										\textcolor{red}{1981年}，卡梅隆执导了第一部作品《食人鱼2》。
										\textcolor{red}{1984年}，卡梅隆推出了他第一部自编自导的影片《终结者》。
										\textcolor{red}{1986年}，自编自导的第二部作品《异形2》问世。
										\textcolor{red}{1987年}，《异形2》获得了第14届土星奖最佳导演奖。
								\textcolor{red}{1991年}，他执导的电影《终结者2》在美国上映后取得了2亿美元的票房。		
									\textcolor{red}{1997年}，执导了电影《泰坦尼克号》。
									\textcolor{red}{1998年}，在第70届奥斯卡金像奖上这部影片获得了14个奥斯卡奖的提名并获得了其中的11个奖项。
				} \\ 
				\bottomrule
			\end{tabular}
				\caption{RQ1: Examples of the generated answers by UTS-\textit{abs}, UTS-\textit{ext} and baselines.}
			\label{tab:case-chi}
		\end{table}
	\end{CJK*}

	 We also show a case study in Table~\ref{tab:case-chi} with translated version in Table~\ref{tab:case-eng} selected from celebrity timelime dataset.
	The case is about James Cameron's career as a director.
	We omit unimportant information in the input document due to limited space.
	The input document includes most of his works, and the detailed information of each event, while the summary reference only introduces the main event of his experience, omitting those details and unimportant events.
	It can be seen that the summary generated by UTS-\textit{abs} successfully captures the important events, and introduces them in the correct order.
	The output of our UTS-\textit{ext} has a high overlap with the ground truth. 
	As for baseline \texttt{GPG}, it fails to capture the most important events, but includes irrelevant information such as details in filming ``Piranha II''.
	For baseline \texttt{SAGCopy}, it also generates unimportant descriptions including information ``The dark angel of the last world''.
	Moreover, our extractive and abstractive summary show consistent behavior with the high overlap, which further indicates that the two methods can jointly promote the recognition of important information. 
	Compared with the extracted summary, the generated summary is more concise and coherent.

	\subsection{Out of Domain Test}
	\label{subsec:tl17}
	Next, we address research question \textbf{RQ2}.
	In Table~\ref{tab:tl17}, we present the performance of UTS on the classic timeline summarization TL17 dataset as an out-of-domain test. 
	It can be seen that both of our models outperform existing baselines.
	Specifically, UTS-\textit{ext} outperforms the best baseline Datawise by 19.4\% on AR1-F score, demonstrating the effectiveness of the neural network in the traditional extractive style.
	UTS-\textit{abs} performs similar to UTS-\textit{ext}, improving the AR1-F score of Pubcount by 3.79.
	This demonstrates that the abstractive methods can be adapted to out-of-domain small-scale datasets.
	Specifically, since our original WikiTS dataset is in encyclopedia style, while Timeline 17 is a news dataset, this demonstrates that our model can be applied to datasets of different language styles.

	\begin{table}[t]
		\centering
		\begin{tabular}{@{}lcc c@{}}
			\toprule
			& AR1-F & AR2-F & Date-F1 \\
			\midrule
			Chieu & 6.66 & 1.9 & 25.1\\
			Martschat & 10.5 & 3.0 & 54.4 \\
			Tran & 9.4 & 2.2 & 51.7 \\
			Pubcount & 10.5 & 2.7 & 48.1 \\
			Datewise & 12.0 & 3.5 & 54.4 \\
			Clust & 8.2 & 2.0 & 40.7 \\
			UTS-\textit{ext} & \textbf{16.73} & \textbf{4.08} & \textbf{54.9} \\
			UTS-\textit{abs} & 14.29 & 3.51 &  54.6\\
			\bottomrule
		\end{tabular}
			\caption{RQ2: ROUGE scores on out-of-domain TL17 summarization dataset.}
		\label{tab:tl17}
	\end{table}

	\subsection{Ablation Study}
	\label{subsec:ablation}
	
	\begin{table}[t]
		\centering
		\begin{tabular}{@{}lcc c@{}}
			\toprule
			& ROUGE-1 & ROUGE-2 & ROUGE-L \\
			\midrule
			UTS-\textit{abs}& 41.56 & 23.95& 39.18\\
			without multitask&39.58 & 22.54 & 37.55 \\
			without global &38.66& 22.87 & 36.76\\
			without local &39.14& 23.15 &  36.00\\
			\midrule
			UTS-\textit{ext}&34.81&22.26&32.03\\
			without multitask & 33.78 & 21.09 & 29.03 \\
			without global &33.00& 18.89 & 27.07\\
			without local &33.28 & 20.98 & 29.69\\	
			\bottomrule
		\end{tabular}
			\caption{RQ3: ROUGE scores of different ablation models.}
		\label{tab:comp_rouge_ablation}
	\end{table}
	
	\begin{figure}
		\centering
		\subfigure{ 
			\includegraphics[clip,width=0.4\columnwidth]{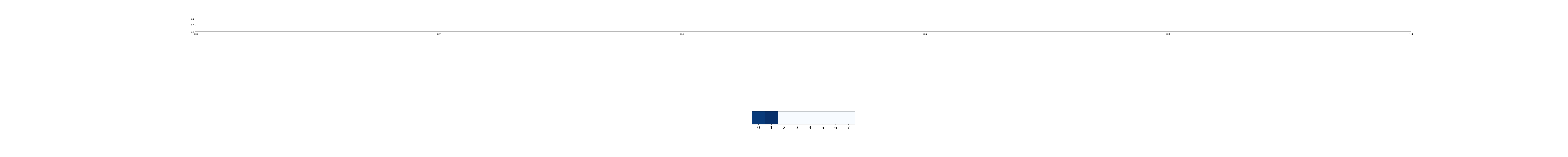}}
		\vspace{-4mm}
		\subfigure{
			\includegraphics[clip,width=0.4\columnwidth]{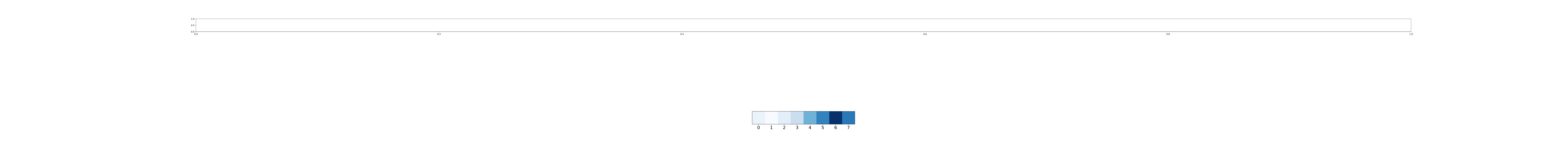}}
		\caption{RQ3: Visualizations of time-attention. The figure in the left part is the attention map in the first decoding step, and the figure in the right part is in the final decoding step.}
		\label{fig:time}
	\end{figure}
	
	Next, we turn to research question \textbf{RQ3}, where we perform an ablation study on the test set to investigate the influence of different modules in our proposed UTS model. 
	Modules are tested in four ways: 
	(1) we remove the sentence extractor and only train the generator to verify the effectiveness of joint learning on the abstractive summarization;
	(2) we remove the summary generator part and only train the sentence extractor to verify the effectiveness of joint learning on the extractive summarization;
	(3) we remove the graph-based encoder and only stores the local representation in the memory to verify the effectiveness of global representation;
	(4) we remove the time-event memory entirely to verify the importance of global and local representation further.

	Table~\ref{tab:comp_rouge_ablation} presents the results. 
	We find that the ROUGE-2 score of extractive summarization drops by 5.26\% after the summary generator is removed. 
	This indicates that the joint learning method helps extractive summarization to benefit from abstractive summarization. 
	ROUGE-2 score of abstractive summarization drops by 5.54\% after the sentence extractor is removed. 
	This indicates that extractive summarization does help abstractive summarization identify important sentences during the interactive decoding phrase.
	ROUGE-2 score of extractive summarization drops by 4.72\%, while the ROUGE-2 score of abstractive summarization drops by 6.25\% after the global representation is removed. 
	It indicates establishing the graph-based encoder to simulate the relationships between events is necessary to improve the performance of both extractive and abstractive summarization. 
	ROUGE-2 score drops by 4.72\% and 3.45\% compared with UTS-\textit{abs} after removing the global representation and the local representation. 
	It indicates the semantic information of the time-event memory is of great importance to encode multiple events.

	\subsection{Analysis of Time Position Embedding}
	\label{subsec:Fidelity}
	
	We then address \textbf{RQ4}. 
	The usefulness of time position embedding is reflected by time-attention in the memory, denoted as $\pi$ as introduced in Equation~\ref{equ:attr-key-score}.
	If the time position embedding successfully encodes the time information, then the time-attention should obey the development of the input document.
	We visualize the attention map of two randomly sampled examples as shown in Figure~\ref{fig:time} from the celebrity timeline dataset.
	The figure on the left is the attention map in the first decoding step, and the figure on the right is in the final decoding step.
	The darker the color is, the higher the attention is. 
	Due to limited space, we omit the corresponding event descriptions.
	When decoding starts, UTS-\textit{abs} learns to pay attention to the first two events, which always consist of parallel information such as the birthplace and birth date of the character.
	The attentions on the last several events are low since it does not need this information in advance.
	When decoding ends, UTS-\textit{abs} focuses more on the last several events.
	However, it also pays attention to the first few events, since timeline summarization is a process of information accumulation, and later sentences should consider previous information.
	The above example demonstrates the effectiveness of time position embedding.

	\begin{figure}
		\centering
		\subfigure{ 
			\includegraphics[clip,width=0.5\columnwidth]{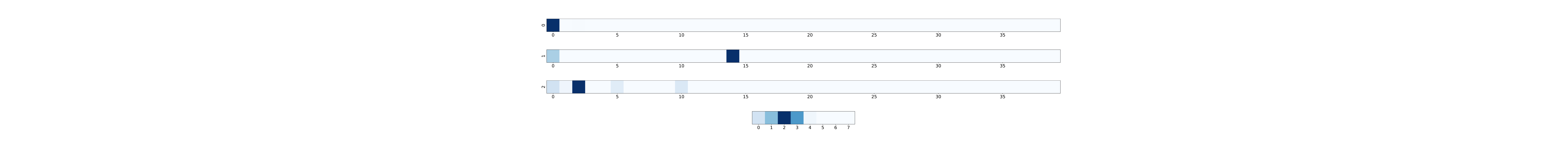}\vspace{-15mm}}
		\vspace{-3mm}
		\subfigure{
			\includegraphics[clip,width=0.5\columnwidth]{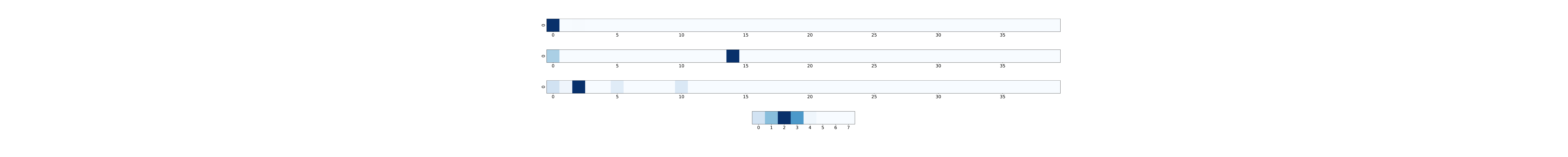}}
		\vspace{-3mm}
		\subfigure{
			\includegraphics[clip,width=0.5\columnwidth]{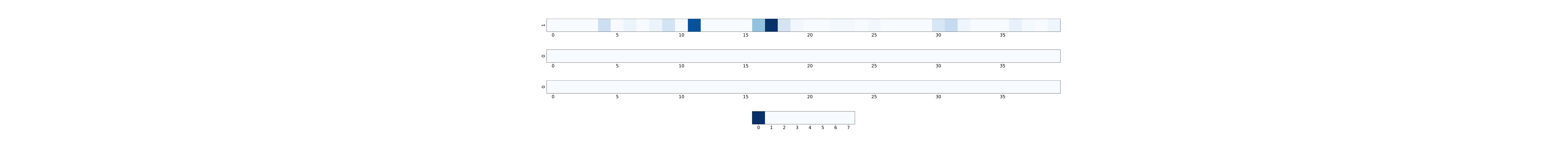}}
		\vspace{-3mm}
		\subfigure{
			\includegraphics[clip,width=0.5\columnwidth]{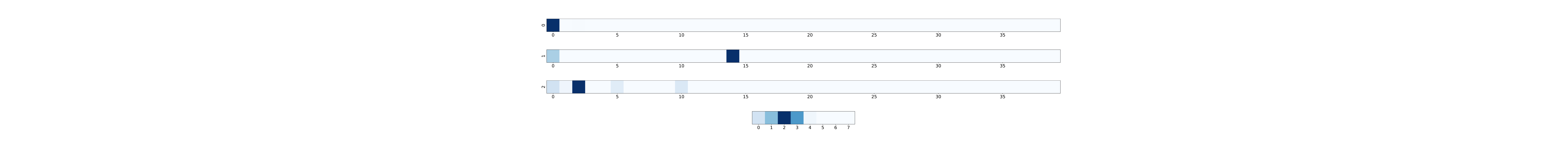}}
		\caption{RQ5: Visualizations of two level attentions. The figure above is the event-level attention and the three figures below are the word-level attentions of first lead three events.}
		\label{fig:two}
	\end{figure}

	\subsection{Analysis of Event-level Attention}
	\label{sec:two}
	We now turn to \textbf{RQ5}, whether event-level attention can guide word-level attention in the abstractive part.
	We first conduct a case study to visualize the two-level attention, as shown in Figure~\ref{fig:two}.
	The figure above is the event-level attention, and the three figures below are word-level attention corresponding to the first three events.
	We only show the first 11 words in an event.
	The result shows that the third event is the most important event in this decoding step, and the weights of the words in this event are also greater than other words on average.
	The above observation demonstrates that event-level attention gives the correct guidance for word-level attention.
	
	Apart from the visualization, we also conduct a quantitative analysis to measure how greatly the word-level attention is influenced by event-level information, which is reflected by inconsistency loss.
	We adjust the inconsistency loss proposed in \S\ref{subsec:unify} to evaluate the inconsistency between event attention and word attention. 
	The new consistency loss at $t$-th decoding step is the negative log-likelihood of the product of attention value of most attended words and their corresponding event-level attention.
	The intuition is to verify whether the event-level attention is high too when word-level attention is high.
	When training starts, the inconsistency loss is around 5.3, and when training ends, the loss drops to 2.1.
	This means that event-level information greatly influences the word-level attention and the model learns to unify these two attentions.
	We did not directly add inconsistency loss to training because we found that made UTS perform worse.
	Instead, we let the model learn by itself to unify these two attentions.
	

	\begin{figure}
	\centering
	\includegraphics[width=0.4\linewidth]{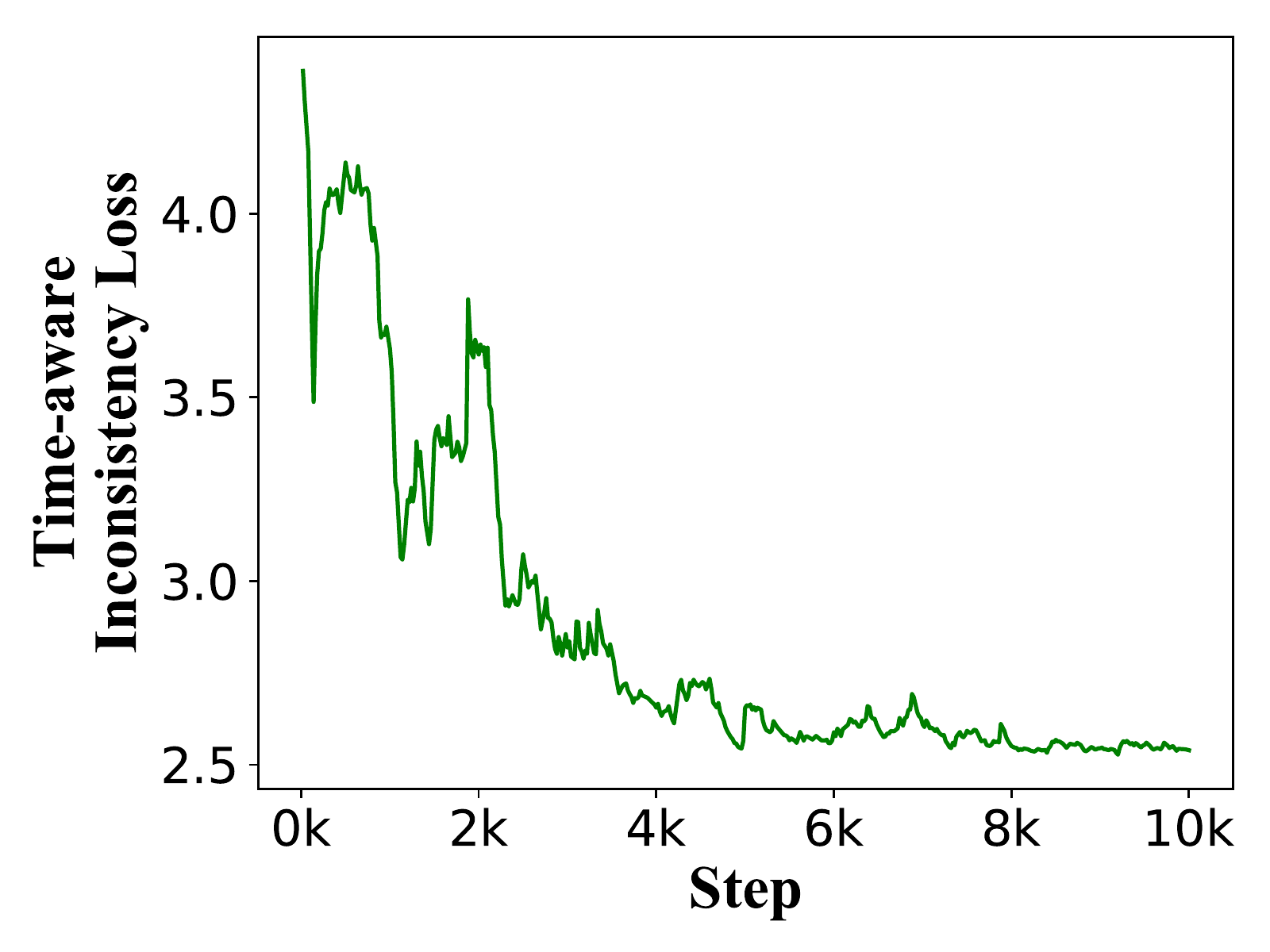}
	\caption{RQ6: Time-aware inconsistency loss curve.}
	\label{fig:consistentloss}
\end{figure}

	\subsection{Analysis of the Unified Chronological Attentions}
	\label{sec:unify}
	
	We then address \textbf{RQ6}, examining whether the chronological attentions in the abstractive and extractive parts are indeed unified.
	Remember that we come up with a time-aware inconsistency loss to unify the two attentions, thus, by looking at the loss curve, we can examine the effectiveness of this part.

	The loss curve of the inconsistency is shown in Figure~\ref{fig:consistentloss}.
	We can see that when the training begins, the inconsistency loss fluctuates from time to time, probably because the model aims to train the extractor and generator separately at the beginning of the process.
	However, the average of the inconsistency loss presents a falling tendency, which means that the extractor and generator unify during the whole training procedure. 
	In the end, the time-aware inconsistency loss drops from 4.0 to 2.5.

    \subsection{Robustness of Parameter Setting}
    \label{robustness}
    
    \begin{figure}[h]
    \centering
    \includegraphics[scale=0.7]{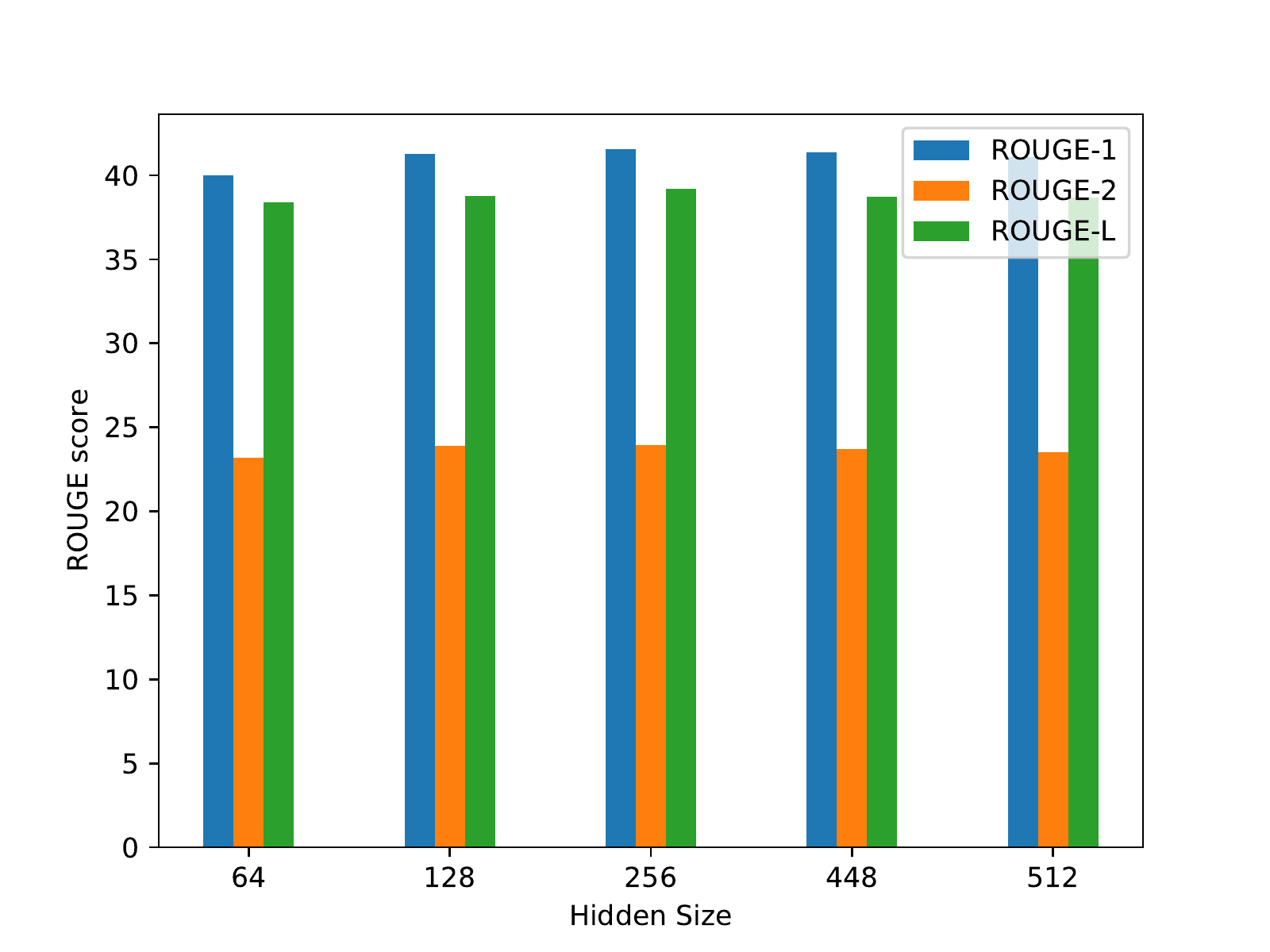}
    \caption{
        Performance of UTS-\textit{abs} with different parameter settings.
    }
    \label{fig:hidden}
\end{figure}

    Finally, we turn to address \textbf{RQ7} to investigate the robustness of parameter setting.
We train our model in different parameter settings as shown in Figure~\ref{fig:hidden}.
The hidden size of the RNN is tuned from 64 to 512, and we use the ROUGE $F_1$ score to evaluate each model.
As the hidden size grows larger from 64 to 256, the performance rises along with.
The increment of hidden size improves the ROUGE-1 and ROUGE-L scores by 0.54 and 0.77 score.
When the hidden size continuously goes larger from 256 to 512, the performance is declined slightly.
The increment of hidden size leads to a 1.15\% and 1.25\% drop in terms of ROUGE-1 and ROUGE-L respectively.
Nonetheless, we can find that each metric is maintained at a stable interval, which demonstrates that our UTS is robust in terms of different parameter sizes.

\section{Conclusion and Future Work}
	\label{sec:conclusion}
	
	In our previous work, we propose a framework named MTS which aims to generate summaries that concisely summarize the evolution trajectory along the timeline.
	However, in this method, the time information is captured in an implicit and indirect way, where it is hard to verify and ensure the decoder indeed captures the time-sequential information.
	Hence, in this work, we propose a novel \emph{Unified Timeline Summarizer} (UTS) that can generate abstractive and extractive timeline summaries in time order.
	Specifically, in the encoder part, we propose a graph-based event encoder that relates multiple events according to their content dependency and learns a representation of each event. 
	In the decoder part, to ensure the chronological order of the abstractive summary, we propose to extract the feature of event-level attention in its generation process with sequential information remained and use it to simulate the evolutionary attention of the ground truth summary.
	The event-level attention can also be used to assist in extracting summary, where we devise a time-aware inconsistency loss function to penalize the inconsistency between abstractive attention and extractive attention.
	Note that the extractive summary is generated one by one, thus the extracted summary also comes in time sequence.
	We augment the character timeline summarization dataset proposed in our previous work with the event timeline summarization corpus and English corpus.
	Experimental results on these datasets and on out-of-domain Timeline 17 dataset show that our UTS model can significantly outperform the existing methods. 
	In the near future, we aim to propose a multi-modal time-aware timeline summarization framework.

\section*{Acknowledgments}
We would like to thank the anonymous reviewers for their constructive comments. 
This work was supported by National Key Research and Development Program of China (No. 2020YFB1406702), National Natural Science Foundation of China (NSFC Grant No. 62122089 \& No. 61876196)

\clearpage
\bibliographystyle{ACM-Reference-Format}
\bibliography{sample-bibliography}

\end{document}